\documentclass[12pt]{article}
\usepackage[colorlinks=true,linkcolor=blue, citecolor=blue]{hyperref}%
\usepackage{geometry}
\usepackage{authblk} 
\usepackage{lineno} 

\usepackage{microtype}
\usepackage{comment}
\usepackage{amsbsy} 
\usepackage{amssymb}
\usepackage{amsmath}
\usepackage[bb=boondox]{mathalfa}
\usepackage[round]{natbib}
\usepackage{csvsimple}
\usepackage{footnote}
\usepackage{tablefootnote}
\usepackage{array}
\usepackage{booktabs}
\usepackage{adjustbox}
\usepackage{multirow}
\usepackage{halloweenmath}
\usepackage{bm}
\usepackage{graphicx}
\usepackage{caption}
\usepackage{subcaption}
\usepackage[justification=justified,singlelinecheck=false]{subcaption}
\usepackage{setspace}
\usepackage{color,soul}

\usepackage[dvipsnames]{xcolor}
\usepackage{float}

\usepackage[normalem]{ulem}

\usepackage{tikz}
\usetikzlibrary{shapes.geometric, arrows}

\begin{document}

\title{Modeling Non-Ergodic Path Effects Using Conditional Generative Model for Fourier Amplitude Spectra}

\author[1,6]{Maxime Lacour \thanks{Corresponding author: maxlacour@berkeley.edu}}

\author[2]{Pu Ren}
\author[3,4]{Rie Nakata}
\author[3,5]{Nori Nakata}
\author[2,6,7]{Michael W. Mahoney}

\affil[1]{\small Department of Civil Engineering, University of California, Berkeley}

\affil[2]{\small Scientific Data Division, Lawrence Berkeley National Lab, Berkeley, CA 94720, USA}
\affil[3]{\small Energy Geosciences Division, Lawrence Berkeley National Lab, Berkeley, CA 94720, USA}
\affil[4]{\small Earthquake Research Institute, University of Tokyo, Tokyo 113-0032, Japan}
\affil[5]{\small Department of Earth, Atmospheric and Planetary Sciences, MIT, Cambridge, MA 02139, USA}

\affil[6]{\small International Computer Science Institute, Berkeley, CA 94704, USA}

\affil[7]{\small Department of Statistics, University of California, Berkeley, CA 94720, USA\vspace{12pt}}

 \date{\today}
\maketitle	

Declaration of Competing Interests

The authors acknowledge there are no conflicts of interest recorded.

\clearpage
\newpage


\maketitle

\clearpage
\newpage


\begin{abstract}

Recent developments in non-ergodic ground-motion models (GMMs) explicitly model systematic spatial variations in source, site, and path effects, reducing standard deviation to 30-40\% of ergodic models and enabling more accurate site-specific seismic hazard analysis. Current non-ergodic GMMs rely on Gaussian Process (GP) methods with prescribed correlation functions and thus have computational limitations for large-scale predictions. This study proposes a deep-learning approach called Conditional Generative Modeling for Fourier Amplitude Spectra (CGM-FAS) as an alternative to GP-based methods for modeling non-ergodic path effects in Fourier Amplitude Spectra (FAS). CGM-FAS uses a Conditional Variational Autoencoder architecture to learn spatial patterns and interfrequency correlation directly from data by using geographical coordinates of earthquakes and stations as conditional variables. Using San Francisco Bay Area earthquake data, we compare CGM-FAS against a recent GP-based GMM for the region and demonstrate consistent predictions of non-ergodic path effects. Additionally, CGM-FAS offers advantages compared to GP-based approaches in learning spatial patterns without prescribed correlation functions, capturing interfrequency correlations, and enabling rapid predictions, generating maps for 10,000 sites across 1,000 frequencies within 10 seconds using a few GB of memory.  
CGM-FAS hyperparameters can be tuned to ensure generated path effects exhibit variability consistent with the GP-based empirical GMM. This work demonstrates a promising direction for efficient non-ergodic ground-motion prediction across multiple frequencies and large spatial domains.

\end{abstract}

\clearpage
\newpage
\section*{Introduction}

Ground-motion models (GMMs) provide the probability distribution of ground-motion intensity measures as functions of earthquake source characteristics, wave propagation path properties, and local site conditions, and are a critical component of probabilistic seismic hazard analysis (PSHA). Traditional ergodic GMMs predict median ground motions based on source, path, and site parameters, but assume that deviations from these median predictions are purely random and do not exhibit systematic spatial patterns \citep{Anderson1999}. This ergodic assumption can lead to inaccurate site-specific hazard estimates because certain regions consistently produce ground motions that are higher or lower than the ergodic median. Non-ergodic GMMs address this limitation by explicitly modeling repeatable spatial variations: systematic source effects at specific earthquake locations, systematic site effects at specific recording stations, and systematic path effects for specific source-to-site pairs. By capturing these systematic spatial patterns, non-ergodic models reduce the standard deviation of ground-motion predictions by 30-40\% compared to ergodic models \citep{Abrahamson2019, Lavrentiadis2023eas}.
Very dense arrays help decompose the source-path-site effects for ground motion modeling, but such arrays are not always available \citep{Nakata2015, Chang2023}. 
In this paper, we focus on generating spatially varying path effects, which are more challenging to model than source and site effects because of the complexity in modeling spatial correlations between source-site pairs. 

The current state-of-the-art approach for non-ergodic GMMs uses Gaussian Processes (GPs), a type of machine learning approaches which rely on prescribed spatial correlation functions between source and site locations, with fixed functional forms \citep{Landwehr2016, Abrahamson2019}. The GP approach has proven effective, but it presents several limitations: the use of prescribed correlation functions may not capture complex spatial patterns in ground-motion data; covariance matrix computations scale with the number of observations and predictions, leading to significant memory requirements for large datasets, even with recent efficiency improvements \citep{Lacour2022}; and a frequency-independent analysis that does not capture interfrequency correlations. 
\cite{Bayless2018b} demonstrated that neglecting interfrequency correlation leads to underestimation of response spectral variability when using random vibration theory, and \cite{Lavrentiadis2023rvt} showed that the uncertainty in pseudo-spectral acceleration increases when interfrequency correlation is considered.


Deep learning approaches have shown promise for modeling ground-motion waveforms and intensities. Early applications of Artificial Neural Networks (ANNs) focused on deterministic predictions of spectral accelerations and peak ground motion parameters using ergodic assumptions. \citep{Kerh2005, Gullu2007, Ahmad2008, Derras2014, Khosravikia2019, Mohammadi2023, Kuran2024, Kang2024}. 
Non-ergodic effects for predicting peak intensity measures have started to be incorporated: \cite{Tani2025} developed a partially non-ergodic model in Greece , while \cite{Cotton2022} introduced a fully non-ergodic convolutional neural network for the Kanto basin. 
 Generative modeling approaches have enabled stochastic predictions of ergodic ground motions that reproduce the observed variability in earthquake recordings. Conditional Generative Adversarial Networks (CGANs) have been used to generate acceleration time histories conditioned on magnitude, distance, and site properties \citep{Florez2022, Esfahani2022}, with \cite{Matsumoto2024} extending this to site-specific predictions in Japan. \cite{Ning2024} combined Conditional Variational Autoencoders (CVAEs) with Gaussian Process regression for spatial interpolation within individual earthquake events. While these generative approaches can generate multiple ground-motion realizations, most also remain ergodic or partially non-ergodic and do not model systematic, repeatable path effects across multiple earthquakes.

Recent developments in Conditional Generative Modeling (CGM) have shown that data-driven generative frameworks can learn complex earthquake source–path–site interactions directly from observations without requiring explicit physical models. CGM ground motion (CGM-GM) uses CVAEs to generate spatially continuous ground-motion time-series by conditioning on earthquake magnitude, depth, and source–receiver coordinates \citep{Ren2024}. CGM-GM learns wave-propagation effects implicitly in the latent space and has been shown to reproduce waveform shapes, P/S arrivals, PGV scaling, and detailed spatial patterns in FAS maps across the San Francisco Bay Area. Building on this, CGM-Wave introduces a diffusion-based generative transformer that produces broadband seismic wavefields, including both amplitude and phase \citep{Bi2025}. Using data from the Geysers geothermal field, CGM-Wave demonstrates high-fidelity reconstruction of waveform moveout, phase coherence, and spatial continuity through a noise-to-wavefield diffusion process and an adaptive phase-retrieval module.

CGM-FAS focuses on generating Fourier Amplitude Spectra (FAS) rather than time-series waveforms like CGM-GM and CGM-Wave. This choice is motivated by the direct relevance of FAS to seismic hazard assessment and the linearity of the Fourier transform, which allows non-ergodic adjustment terms to be transferred from smaller to larger magnitude earthquakes—a property not shared by peak measures such as spectral acceleration \citep{Goulet2015, Lavrentiadis2023eas}. 
We implement CGM-FAS using a CVAE because of its lightweight performance, ease of training, and robustness against mode collapse issues common in CGANs. 
The approach offers several advantages: capturing spatial continuity without prescribed correlation functions, learning interfrequency dependencies directly from data, and providing computationally efficient alternatives to GP methods. 
We evaluate CGM-FAS using a small-magnitude earthquake dataset from the San Francisco Bay Area \citep{Lacour2025c}, downloaded from the NCEDC database \citep{ncedc}, with station and earthquake locations shown in Figure \ref{fig:Stations_and_Sources}. The evaluation compares CGM-FAS with GP-based approaches across spatial correlation patterns, interfrequency correlations, computational efficiency, and uncertainty quantification. Mathematical notations are described in Table~\ref{tab:sup_math_notation}.


\section*{Non-Ergodic Path Effects}

Following \cite{Lacour2025b}, we describe a non-ergodic GMM for velocity FAS of small earthquakes in the San Francisco Bay Area. We refer to this non-ergodic model as the LANN25 model. 
The model takes the form:
\begin{equation} \label{eq:NonergGMM}
\begin{aligned}
\ln FAS_{NE}(M,R_{Rup}, Z_{hyp}, {te}_{e},{ts}_s) = 
   & \ln FAS_{E}(M,R_{Rup}, Z_{hyp}) + \\
   & \delta L2L_e({te}_{e})+\delta S2S_s({ts}_s) + \delta P2P_{es}({te}_{e},{ts}_s) + \\
   & \delta B_e^0(t_e) + \delta WSP_{es}(t_e, t_s) , 
\end{aligned}
\end{equation}
where $\ln FAS_{NE}$ is the non-ergodic GMM, $M$ is the magnitude, $R_{Rup}$ is the rupture distance in km, $Z_{hyp}$ is the hypocentral depth in km, and ${te}_{e}$ and ${ts}_{s}$ are the coordinates of the earthquake $e$ and site $s$, respectively. The ergodic-base GMM $\ln FAS_{E}$ includes the magnitude, rupture distance, and depth scalings, whose forms and coefficients can be found in \cite{Lacour2025b}.  The ergodic-base model does not include velocity parameters such as a $V_{S30}$ because of a lack of their reliable estimates in this area \citep{cpr_vm1}.
Excluding $V_{S30}$ from the ergodic model formulation implicitly incorporates its effect in the non-ergodic model and does not reduce the model applicability. The terms $\delta L2L_e$, $\delta S2S_s$ and $\delta P2P_{es}$ are the mean deviation from the ergodic model for the source, site, and path terms, respectively, and represent their spatial patterns. 
The terms $\delta B_e^0$ and $\delta WSP_{es}$ are the remaining residuals of the source and within-path components, respectively. These residuals capture the inherent, unpredictable event-to-event randomness in ground motions, and therefore are referred to as aleatory residuals.

We focus on estimating the mean path term $\delta P2P_{es}$ from within-site residuals, which we refer to as the path term throughout the rest of this paper. 
To estimate the path terms, we first estimated within-site residuals by removing the ergodic-base model and the spatially varying coefficients from the observations in Equation \eqref{eq:NonergGMM}. These within-site residuals can be decomposed into two components: 
\begin{equation}\label{eq:within_site_residuals_1}
    \delta WS_{es}(te_{e},ts_s) = \delta P2P_{es}(te_{e},ts_s) + \delta WSP_{es}({te}_{e},{ts}_{s}).
\end{equation}

The term $\delta P2P_{es}$ is obtained as the mean of the within-site residuals.
The term $\delta WSP_{es}$ is computed by removing the path term $\delta P2P_{es}$ from the within-site residual $\delta WS_{es}$, and expected to be spatially uncorrelated with zero mean and standard deviation $\phi_{SP, NE}$, which is refereed to as aleatory variability.

\subsection*{Gaussian Process}

\newcommand{\dPtoP}{\mathbf{\delta P2P}}
\newcommand{\dPtoPpred}{\mathbf{\boldsymbol{\boldsymbol{\delta P2P^{(f)}_\mathrm{pred}}}}}

\newcommand{\dWS}{\mathbf{\boldsymbol{\delta WS}}}
\newcommand{\dWSobs}{\mathbf{\boldsymbol{\delta WS_\mathrm{obs}}}}
\newcommand{\dWSpred}{\mathbf{\boldsymbol{\delta WS_\mathrm{pred}}}}
\newcommand{\Koo}{\mathbf{\boldsymbol{K_\mathrm{obs,obs}}}}
\newcommand{\Kop}{\mathbf{\boldsymbol{K_\mathrm{obs,pred}}}}
\newcommand{\Kpp}{\mathbf{\boldsymbol{K_\mathrm{pred,pred}}}}
\newcommand{\vtheta}{\boldsymbol{\theta}}
\newcommand{\robs}{\mathrm{obs}}
\newcommand{\rest}{\mathrm{est}}

The LANN25 model uses GPs to learn path effects from within-site residuals by defining a spatial correlation function that quantifies how similar the ray paths are between different source-site pairs \citep{Sung2023, Liu2025,Lacour2025b}. The LAAN25 model estimates path effects independently per frequency and assumes all the terms as frequency-dependent while no interfrequency correlation is included. Hereafter, we denote the within-path residuals for multiple source-site pairs as $\dWS$. The posterior distribution of the within-site residuals at a given frequency conditioned on the observations is assumed to follow a Gaussian distribution: 

\begin{equation}
    p_{\boldsymbol{\theta_{GP}}}(\dWSpred  | \dWSobs, \boldsymbol{\mathbf{te}_{\mathrm{obs}}},  \boldsymbol{\mathbf{ts}_{\mathrm{obs}}}, \
    \boldsymbol{\mathbf{te}_{\mathrm{pred}}},  \boldsymbol{\mathbf{ts}_{\mathrm{pred}}}) 
    =
   \mathcal{N}(\boldsymbol{\dPtoP}_{\boldsymbol{\mathrm{pred}},\vtheta_{\boldsymbol{\mathrm{GP}}}},  \boldsymbol{\psi}_{\boldsymbol{\mathrm{pred}},\vtheta_{\boldsymbol{\mathrm{GP}}}}^2), \label{eq:pdf_GP}
\end{equation} 
where $\boldsymbol{\mathbf{te}_{\mathrm{obs}}}$ and $\boldsymbol{\mathbf{te}_{\mathrm{pred}}}$ are the locations of earthquakes,
$\mathbf{ts}_{\mathrm{obs}}$ and $\mathbf{ts}_{\mathrm{pred}}$ are the locations of sensors for the observed and target scenarios, respectively, $\boldsymbol{\dPtoP}_{\boldsymbol{\mathrm{pred}},\vtheta_{\boldsymbol{\mathrm{GP}}}}$ and  $\boldsymbol{\psi}_{\boldsymbol{\mathrm{pred}},\vtheta_{\boldsymbol{\mathrm{GP}}}} $ are the path effects and the corresponding epistemic uncertainty at the predicted earthquake and station locations, and $\vtheta_{\boldsymbol{\mathrm{GP}}}$ is the hyperparameter of the GP.

The predicted path effects and epistemic uncertainty from the Gaussian distribution in \eqref{eq:pdf_GP} are given by: 
\begin{equation} 
    \boldsymbol{\dPtoP}_{\boldsymbol{\mathrm{pred}},\vtheta_{\boldsymbol{\mathrm{GP}}}} = {\Kop}^\mathrm{T} (
        \Koo  + \phi_{SP, NE}^2  \mathbf{I}_d)^{-1}  \dWSobs \label{eq:mu_path},
\end{equation}
\begin{equation}  
    \boldsymbol{\psi}_{\boldsymbol{\mathrm{pred}},\vtheta_{\boldsymbol{\mathrm{GP}}}} = 
    \text{diag}(\Kpp -  {\Kop}^\mathrm{T} ( \Koo  + \phi_{SP, NE}^2  \mathbf{I}_d)^{-1} \Kop) \label{eq:psi_path}  .
\end{equation}
The covariance matrices $\Koo$, $\Kpp$, and $\Kop$ represent the prior spatial correlations of path effects between observed source-site pairs, predicted source-site pairs, and between observed and predicted pairs, respectively. The functional form of the covariance between paths $P$ and $P'$ is given by:

\begin{equation} \label{eq:FAS_correlation_path}
    k_{PP'} = \phi_{P2P}^2 \: \exp\left( \dfrac{-\Delta R_{rup}^2}{2\rho_{R}^2}\right) \exp\left( \dfrac{-\Delta Az^2}{2\rho_{Az}^2}\right )  \exp\left(\dfrac{-|SS'|^2}{2\rho_{S}^2}\right )   + \phi_{SP,NE}^2 \ \delta(P, P') ,
\end{equation}
where $|SS'|$ is the distance between site locations, $\Delta Az$ is the difference in azimuths, $\Delta R_{rup}$ is the difference in rupture distances, and $\delta (P, P') = 1$ if the paths are identical, and 0 otherwise. The relationships between $|SS'|$,  $\Delta Az$  and $\Delta R_{rup}$ are illustrated in \cite{Lacour2025a}.
The standard deviations and correlation lengths are the hyperparameters $\vtheta_{\boldsymbol{\mathrm{GP}}}$. These include $\phi_{P2P}$, which represents the standard deviation of the spatially correlated path effects, $\phi_{SP,NE}$, which represents the standard deviation of the within-path residuals which cannot be explained by the correlation function, and $\rho_R$, $\rho_{Az}$ and $\rho_S$ which represent the correlation lengths of the rupture distance, azimuth, and site separation distance, respectively.

During the training stage, the correlation lengths, $\rho_{R}$, $\rho_{Az}$, and $\rho_{S}$, were estimated by fitting the semi-variogram of the within-site residuals at each frequency prior to conducting the regression. With the correlation lengths held fixed, the variance terms  $\phi_P$ and $\phi_{SP, NE}$ were estimated by maximizing the log-likelihood function, which is defined as:
%
\begin{align}
\log p_{\boldsymbol{\theta_{GP}}}(\dWSobs \mid \vtheta_{\boldsymbol{\mathrm{GP}}}) 
&= -\tfrac{1}{2} 
(\dWSobs)^\mathrm{T}
\left( \Koo + \phi_{SP, NE}^2 \mathbf{I}_d \right)^{-1}
(\dWSobs) \nonumber \\
&\quad - \tfrac{1}{2} 
\log \left| \Koo + \phi_{SP, NE}^2 \mathbf{I}_d \right|
- \dfrac{n_{obs}}{2} \log(2\pi),
\label{eq:loglik_gp}
\end{align}

where $n_{obs}$ is the size of the observed dataset.  The estimates of the hyperparameters are listed in \cite{Lacour2025b}.

To predict path effects at new earthquake and station locations, the GP requires computing and inverting a large $n_{\mathrm{obs}} x n_{\mathrm{obs}}$ covariance matrix in Equations \eqref{eq:mu_path} and \eqref{eq:psi_path}. The computational time scales as $O(n_{\mathrm{obs}}^3)$ and memory usage scales as $O(n_{\mathrm{obs}}^2)$, where $n_{\mathrm{obs}}$ is the number of training observations.These computational costs make simultaneous multi-frequency predictions and large-scale applications (e.g., dense gridded station networks) computationally challenging.

\subsection*{Conditional Variational Autoencoder}

\newcommand{\rdecoder}{\mathrm{decoder}}
\newcommand{\rencoder}{\mathrm{encoder}}

We develop CGM-FAS based on CVAE for velocity FAS, following CGM-GM of \cite{Ren2024} that focuses on generating time series. 
Instead of directly estimating path effects as in a GP, CGM-FAS takes an alternative approach by learning the posterior distribution of the within-site residuals. From this learned distribution, the path effects $\boldsymbol{\dPtoP}$ between multiple source-site pairs can then be estimated.  The computational efficiency of CVAE allows us to estimate the distribution over multiple frequencies, and  $\dWS$ in this section represents the within-site residuals for multiple source-station pairs across frequencies.

CGM-FAS consists of three main components: an encoder that compresses observed within-site residuals into a latent representation, a decoder that reconstructs within-site residuals from this latent representation, and an embedding module that incorporates earthquake and station location information into both the encoder and decoder. 
The posterior distribution of the within-site residuals $p_{\boldsymbol{\mathbf{\theta}_{\rdecoder}}}(\dWS|\dWSobs, \mathbf{te}, \mathbf{ts})$ is parameterized by the decoder hyperparameters $\boldsymbol{\theta}_{\rdecoder}$ as follows:
\begin{equation} \label{eq:pdf_decoder}
p_{\boldsymbol{\mathbf{\theta}_{\rdecoder}}}(\dWS|\dWSobs, \mathbf{te}, \mathbf{ts}) = \int p_{\boldsymbol{\mathbf{\theta_{\rdecoder}}}}(\dWS |\mathbf{te}, \mathbf{ts}, \mathbf{z}) p_{\boldsymbol{\mathbf{\theta_{\rdecoder}}}}(\mathbf{z}) \mathbf{dz}, 
\end{equation}
where $\mathbf{z}$ is a latent variable that provides a compact representation of the within-site residuals.  We assume that $p_{\boldsymbol{\mathbf{{\theta_{decoder}}}(\mathbf{z})}}$ follows a Gaussian distribution $\mathbf{z} \sim  \mathcal{N}( \boldsymbol{0},  \mathbf{{I})}$ with probability distribution $p(\mathbf{z})$. To make the generations meaningful, the latent space needs to represent the observations. This is achieved by introducing an encoder $p_{\boldsymbol{\mathbf{{\theta_{encoder}}}}}(\mathbf{z}|\dWSobs, \mathbf{te}, \mathbf{ts} )$, parameterized by $\boldsymbol{\mathbf{\theta_{encoder}}}$, which are the hyperparameters of the encoder network. 

During the training stage, we optimize $\boldsymbol{\theta_{\rencoder}}$ and $\boldsymbol{\theta_{\rdecoder}}$ using the encoder and decoder. 
For the given observations $\dWSobs$, we aim to minimize the following loss function ~\citep{kingma2013auto} 
\begin{equation}\label{eq:loglik_CGMFAS}
\begin{split}
   \mathcal{L}(\boldsymbol{\theta}_{\mathrm{encoder}},\boldsymbol{\theta}_{\mathrm{decoder}};\boldsymbol{\delta WS}_{\mathrm{obs}}) 
    =& \text{MSE}(\mathbf{\dWS_{obs}}, \hat{\mathbf{\dWS_{obs}}}) + \\
    & \alpha \cdot D_{KL}(p_{\boldsymbol{\mathbf{\theta}_{\rencoder}}}(\mathbf{z}|\mathbf{\dWS_{obs}}) ||  p(\mathbf{z})),
    \end{split}
\end{equation}
where MSE is the mean-squared error operator between $\mathbf{\dWS_{obs}}$ and the reconstructed $\hat{\mathbf{\dWS_{obs}}}$ using CGM-FAS,  $D_{KL}(\cdot)$ denotes the Kullback-Leibler (KL) divergence between the approximate and true posterior distributions, and $\alpha$ is a parameter that gives a balance between the reconstruction loss and the KL divergence. 
Equation \eqref{eq:loglik_CGMFAS} can be thought of as the equivalent of Equation \eqref{eq:loglik_gp} to the GP approach when optimizing the hyperparameters. The number of hyperparameters $\boldsymbol{\theta_{\rencoder}}$ and $\boldsymbol{\theta_{\rdecoder}}$ is much larger for CGM-FAS, depending on the number of layers, than the hyperparameters for GP $\vtheta_{\boldsymbol{\mathrm{GP}}}$. After training, we estimate the path term for each source-site pair $\delta P2P_\mathbf{{\mathrm{pred},VAE}}({te}_{e},{ts}_s)$ by generating additional samples of $\dWSpred({te}_{e},{ts}_s)$ and taking their average.

\begin{figure}[H]
\centering
         \includegraphics[width = 1.2\columnwidth]{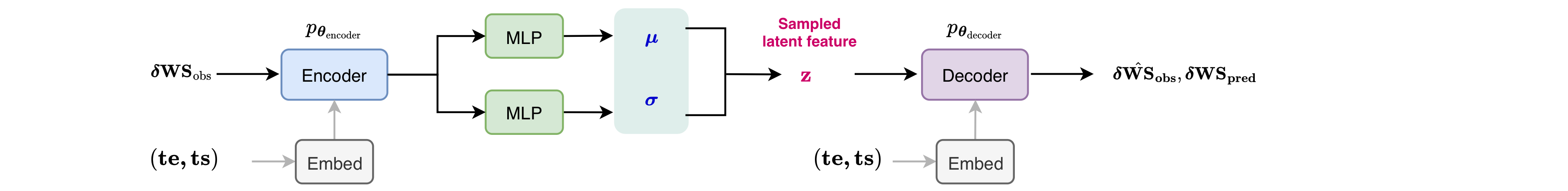} 
     \caption{Illustration of the architecture of CGM-FAS. The encoder $p_{\boldsymbol{\mathbf{\theta}_{\rencoder}}}$ uses convolutional layers to extract  features from observed within-site residuals ${\mathbf{\dWS_{obs}}}$ conditioned on event and site coordinates ($\mathbf{te}$, $\mathbf{ts}$) and map them to a low-dimensional latent space . A Multi-Layer Perceptron (MLP) layer outputs output the parameters of a latent distribution: mean $\boldsymbol{\mu}$ and standard deviation $\boldsymbol{\sigma}$. A latent variable $\mathbf{z} \sim \mathcal{N}( \boldsymbol{0},  \mathbf{{I})}$ is sampled from a standard normal distribution and passed through the decoder $p_{\boldsymbol{\mathbf{\theta}_{\rdecoder}}}$ to generate the reconstructed observation $\hat{\mathbf{\dWS_{obs}}}$ conditioned on the same event and site coordinates. Here, $\boldsymbol{\theta_{\rencoder}}$ and $\boldsymbol{\theta_{\rdecoder}}$ represent the hyperparameters of the encoder and decoder neural networks, respectively.
     } 
     \label{fig:CGM-FAS_Architecture}
\end{figure}

We implement CGM-FAS using the architecture illustrated in Figure \ref{fig:CGM-FAS_Architecture}. During training, the encoder uses convolutional layers to extract FAS features from observed within-site residuals and map them to a low-dimensional latent space. A Multi-Layer Perceptron (MLP) layer outputs the mean $\boldsymbol{\mu}$ and standard deviation $\boldsymbol{\sigma}$ of the latent Gaussian distribution. The decoder samples the latent variable $\mathbf{z}$ from this distribution and uses deconvolutional layers to progressively reconstruct the within-site residuals across all frequencies. Conditioning parameters ($\mathbf{te}$, $\mathbf{ts}$) are embedded by MLP layers and fed into both encoder and decoder for location-specific predictions. During inference, $\mathbf{z}$ is sampled directly from $\mathcal{N}(\mathbf{0}, \mathbf{I})$ and passed through the decoder to generate within-site residuals.


\section*{Data, Training, and Tuning}

The dataset of small-magnitude earthquakes in the San Francisco Bay Area region compiled by \cite{Lacour2025c} and used by LANN25 is used. We extracted the East-West components for the study. and compared with the performance of East-West component predictions of LANN25. This dataset includes stations within a rectangular area placed 50~km away from the edges of the Hayward fault and earthquakes recorded within 100 km of the stations. Noisy data were removed by applying a minimum signal-to-noise ratio of 3 for all frequencies across the 2-15 Hz frequency band.  
The final dataset includes 5,108 recordings from 266 stations and 737 events with magnitudes between 2 and 4. 
The non-ergodic path effects were modeled from the within-site residuals of at 911 available frequencies.

The CGM-FAS architecture and training procedure were determined through systematic experimentation. We gradually increased the number of layers and channels in the encoder and decoder while minimizing the loss function using an initial value of $\alpha = 0.0001$ in Equation \eqref{eq:loglik_CGMFAS}. We found that 3 layers with 32 channels produced smooth reconstructions of the within-site residuals through visual inspection. We then tuned the trade-off parameter $\alpha$  to further reduce the loss function. We observed that the range of generated data was sensitive to $\alpha$, with larger values of $\alpha$ producing narrower ranges. We selected $\alpha$ so that the standard deviation of the within-path residuals $\phi_{SP,NE}$ from CGM-FAS predictions over the available dataset matches the value of 0.40. 
The value of $\phi_{SP, NE}$ has  been estimated only recently using empirical non-ergodic GMMs, and its value has been shown to be approximately the same for different regions around the world and at different frequencies, with a value of 0.40 \citep{Sung_Abrahamson_2024b} including the LANN25 model. 
A more detailed description of the calibration procedure for $\alpha$ and its effect on the range of the generated data is provided in the Supplementary Material Section \ref{section:Supplement_Alpha}.

After training, CGM-FAS can generate predictions of the within-site residuals at arbitrary earthquake and station coordinates by prescribing new conditioning parameters $\mathbf{te}$ and $\mathbf{ts}$. We obtain the path effect of a specific source-site pair by generating 200 generations and taking their average. This number was chosen to ensure convergence of both the mean and standard deviation of the predicted within-site residuals, as demonstrated in the Supplementary Section \ref{section:Supplement_Convergence}. A single training of CGM-FAS required approximately 2 hours on a single Apple M1 CPU. An inference of a single source-site pair over the frequency range took 1e-5 sec. The trained model contains 678,000 parameters and requires 2 GB of storage.

\section*{Results}

\subsection*{Within-Path Residuals}

\begin{figure}[H]
\centering
\begin{subfigure}[b]{0.45\linewidth}
        \centering
        \includegraphics[width=\linewidth]{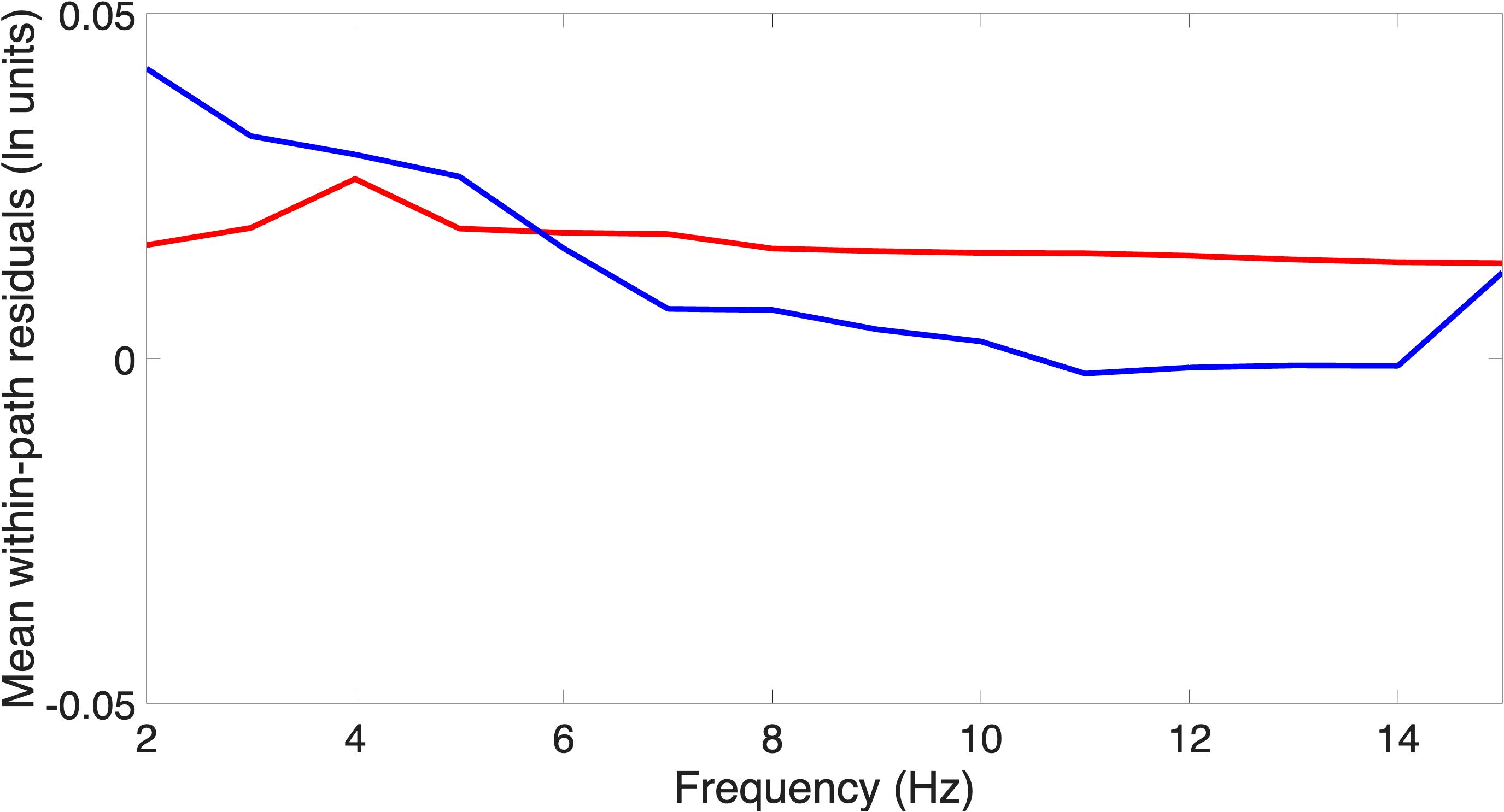}
        \caption{}
        \label{fig:mean_SP_VAE_GMM}
    \end{subfigure}
    \hfill
 \begin{subfigure}[b]{0.45\linewidth}
        \centering
        \includegraphics[width=\linewidth]{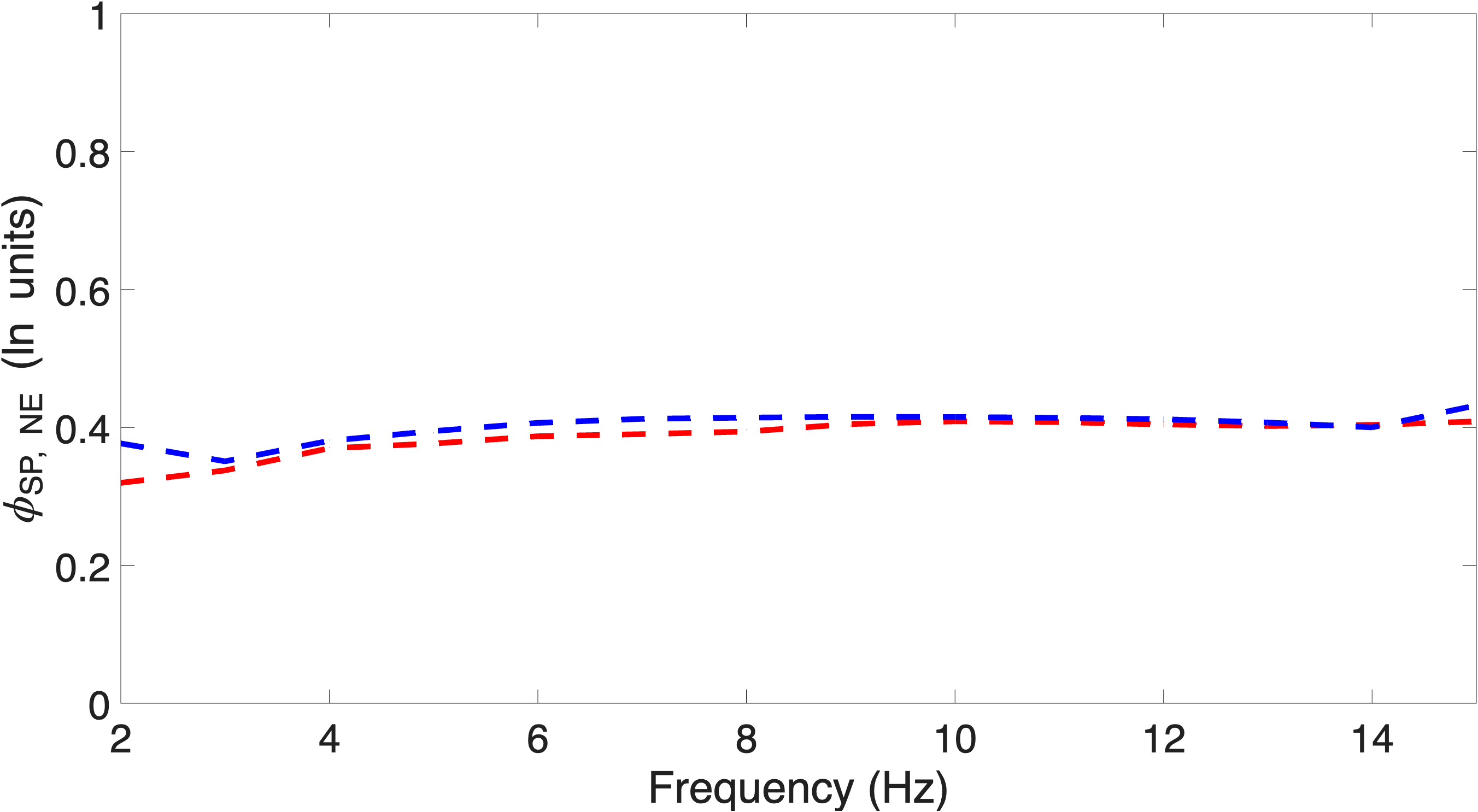}
        \caption{}
        \label{fig:phi_SP_VAE_GMM}
    \end{subfigure}
\hfill
\begin{subfigure}{0.45\linewidth}
    \centering
    \includegraphics[width=\linewidth]{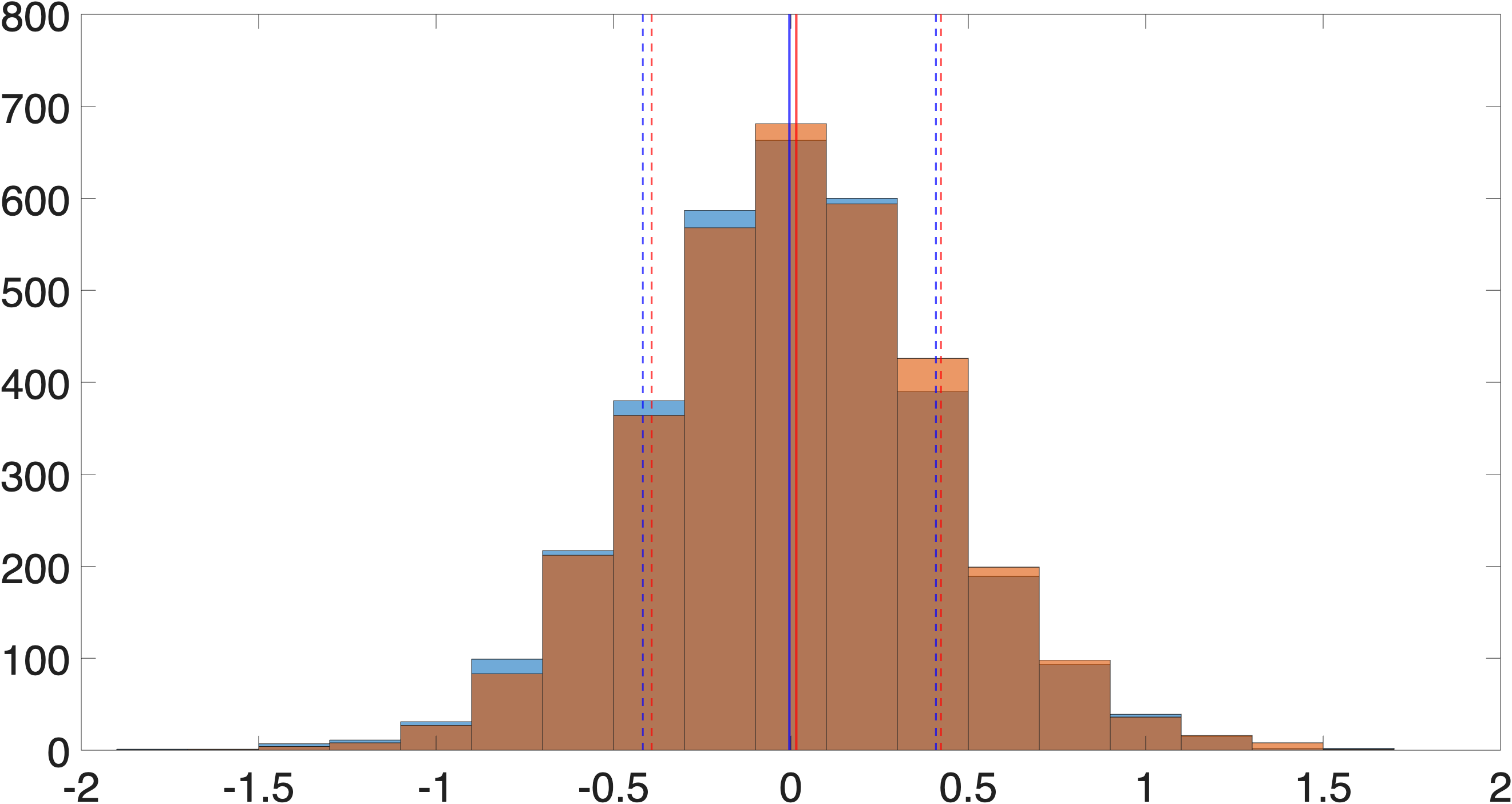}
        \caption{} \label{fig:Histogram_WSP_GMM_VAE_10Hz}
\end{subfigure}

\caption{Comparison of within-path residuals from LANN25 (red) and CGM-FAS (blue). (a) Mean residuals versus frequency, (b) standard deviation $\phi_{SP,NE}$ versus frequency, (c) histograms at 10 Hz with mean (solid lines) and standard deviation (dashed lines).  Note that $\phi_{SP,NE}$ of the GP in (b) is computed from the within-path residuals at the available stations rather then the value estimated using Equation \eqref{eq:FAS_correlation_path}.}
\label{fig:validation_residuals}
\end{figure}

To validate the performance of CGM-FAS on the available dataset, we predicted path effects at all available source-site pairs and evaluated the within-path residuals by subtracting them from the observed within-site residuals as in Equation \eqref{eq:within_site_residuals_1}.  
Figures \ref{fig:mean_SP_VAE_GMM} and \ref{fig:phi_SP_VAE_GMM} show the mean and the standard deviation of the within-path residuals obtained from CGM-FAS and LANN25 across all available source-site pairs in the dataset.  Figure \ref{fig:Histogram_WSP_GMM_VAE_10Hz} shows the histograms of the within-path residuals at 10 Hz for LANN25 and CGM-FAS.

Both methods produce the within-path residuals that approximately follow a Gaussian distribution centered near zero as seen in the histograms, demonstrating that no bias is present in the predictions from both models. LANN25 shows mean residuals approximately 0.01 to 0.02 ln units across frequencies, while CGM-FAS shows mean residuals of 0.04 ln units at 2 Hz that decreases with frequencies and are smaller than LANN25 above 6 Hz. These mean residual values little affect ground-motion predictions and can be considered negligible for seismic hazard applications. The standard deviation of within-path residuals is nearly identical between both methods across the entire frequency range, due to the way we tuned CGM-FAS. LANN25 and CGM-FAS maintain a relatively constant standard deviation of approximately 0.40 ln units across the entire frequency range. They deviate at low frequencies (2-4 Hz), where the values from LANN25 decrease. These results demonstrate that CGM-FAS achieves comparable performance to the LANN25 approach in path term predictions where observations are available.

\subsection*{Spatial interpolation and continuity} \label{section:Spatial_Interpolation}

\begin{figure}[H]
\centering
    \begin{subfigure}[b]{0.45\linewidth}
        \centering
        \includegraphics[width=\linewidth]{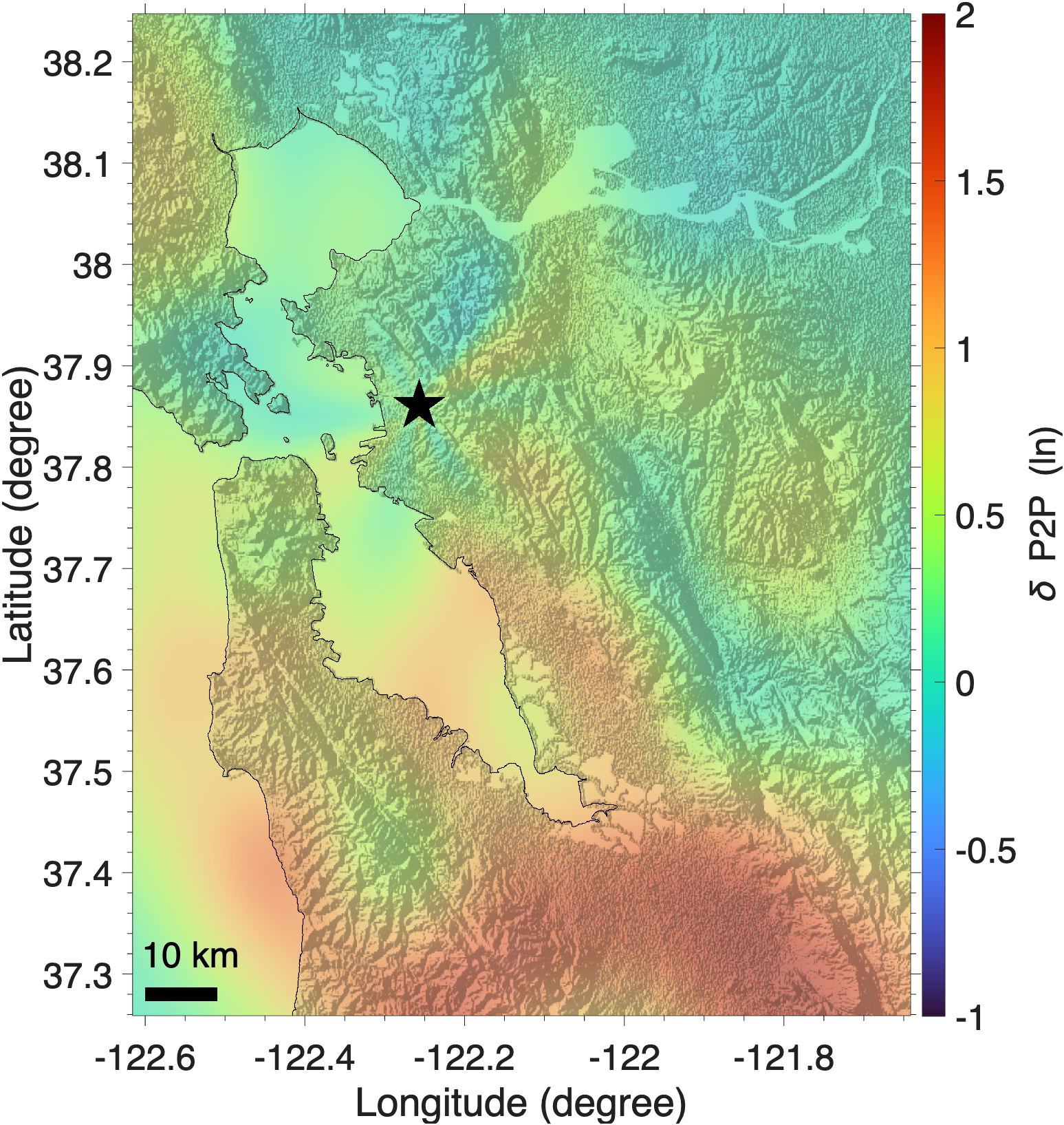}
        \caption{}\label{fig:Map_LANN25}
    \end{subfigure}
    \hfill
    \begin{subfigure}[b]{0.45\linewidth}
        \centering
        \includegraphics[width=\linewidth]{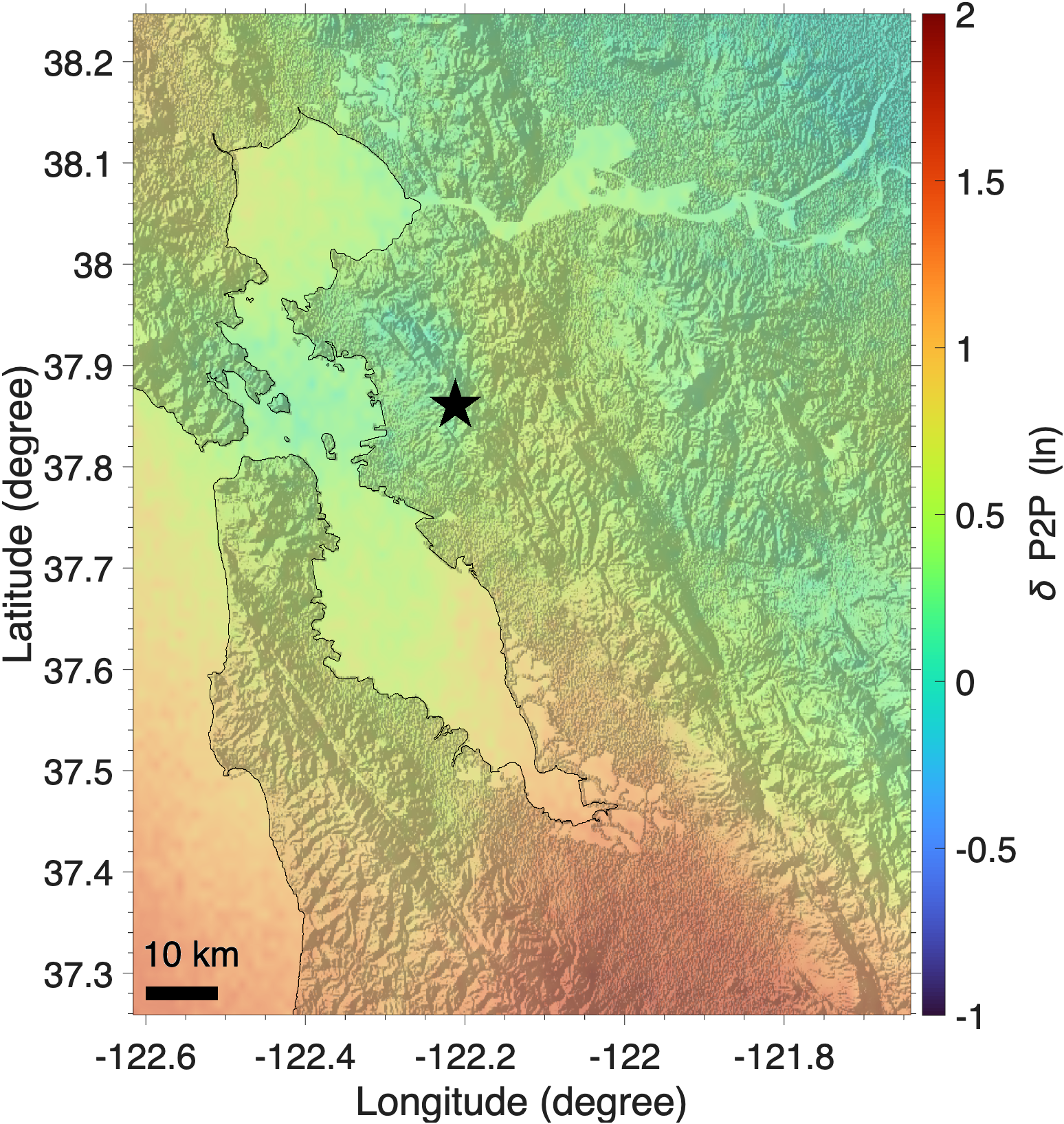}        
        \caption{}\label{fig:Map_CGM}
    \end{subfigure}
    
    \vspace{0.5cm} 
    
    \begin{subfigure}[b]{0.45\linewidth}
        \centering
        \includegraphics[width=\linewidth]{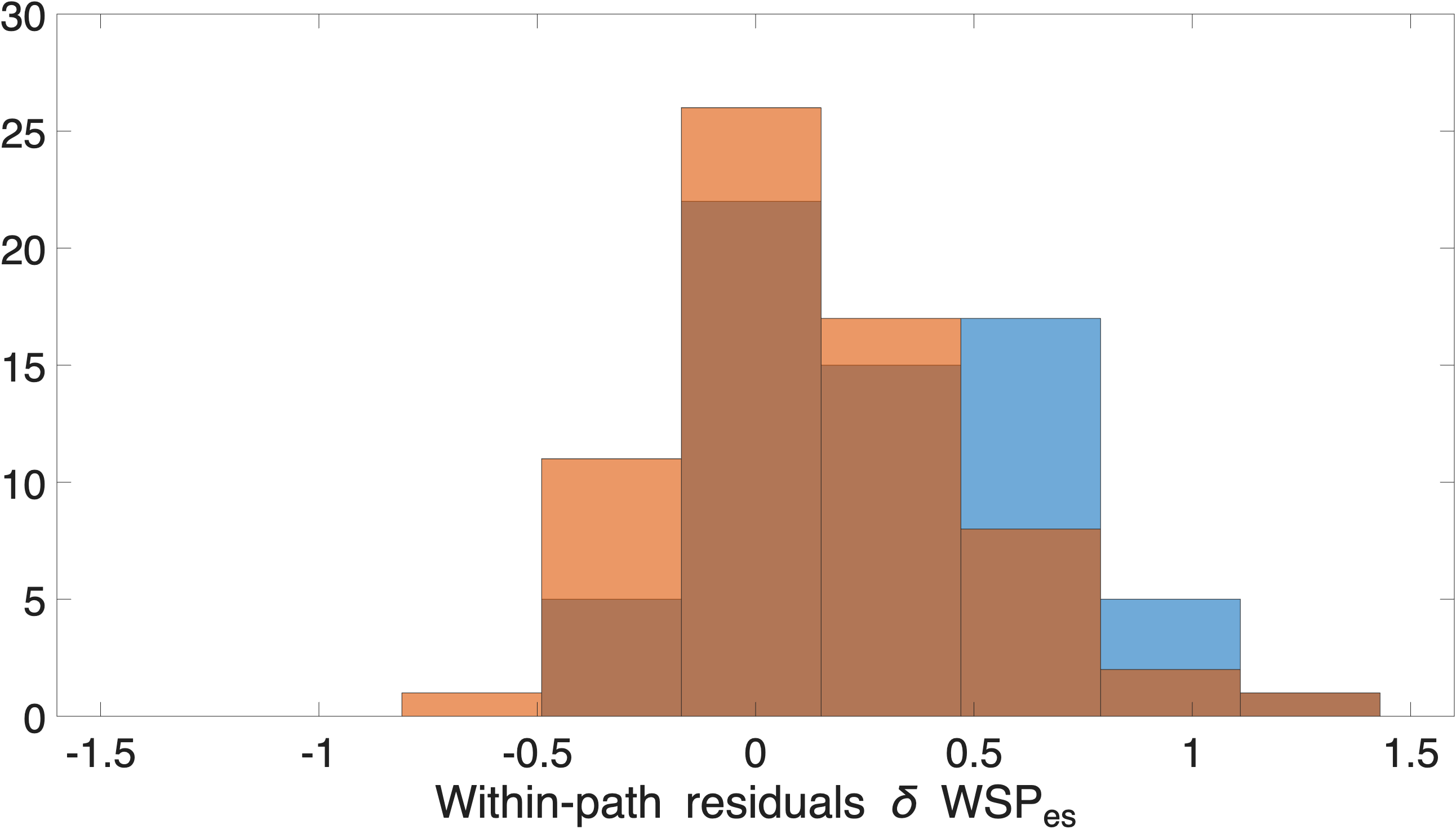}
        \caption{} \label{fig:Histogram_WP_residuals_Berkeley}
    \end{subfigure}
\caption{Comparison of path effects prediction between LANN25 (a) and CGM-FAS (b) at 10 Hz. (ec Histogram of within-path residuals from LANN25 (red) and CGM-FAS (blue). }
 \label{fig:Maps_Path_Pred_GP_VAE}
\end{figure}

We illustrate the capabilities of CGM-FAS to predict path terms where stations are not available, and  compare our predictions with LANN25. As an example, we select a scenario of the earthquake that occurred in 2011 in Berkeley, California, of magnitude $M_w = 3.8$ located at a depth of $Z = 8$ km on the Hayward fault. 
We performed a prediction of the path effects over sites located on a 1 km-spaced square grid inside the domain indicated by the black box in Figure \ref{fig:Stations_and_Sources}.
The map of the predictions of path effects at 10 Hz from LANN25 and CGM-FAS are shown along with the within-path residuals and their histograms in Figure \ref{fig:Map_LANN25} and \ref{fig:Map_CGM} respectively. The within-path residuals are computed by subtracting the predicted path effects from the observed within-site residuals at 66 stations that recorded the earthquake. The histograms of the within-path residuals from the two methods are shown in Figure \ref{fig:Histogram_WP_residuals_Berkeley}.

Both CGM-FAS and LANN25 predict similar continuous spatial patterns, and the within-path residuals show no extreme outliers or long tails, which confirms their good performance. Spatial variations in the path effects indicate localized deviations from the assumed velocity or attenuation structure \citep{Sung2023}, and our predictions are consistent with known structural variations in the region \citep{Aagaard2008, Hirakawa2022}. Both models show large path effects in the South Bay, which may suggest larger-than-average ground motions in areas underlain by the shallow basement and relatively fast, low attenuation materials west of the Hayward Fault. In contrast, smaller path effects east of the Hayward Fault align with the presence of deeper, slower, and more attenuative sediments.

\begin{figure}[H]
    \centering
    
  \begin{subfigure}{.45\linewidth}
        \centering
        \includegraphics[width=\linewidth]{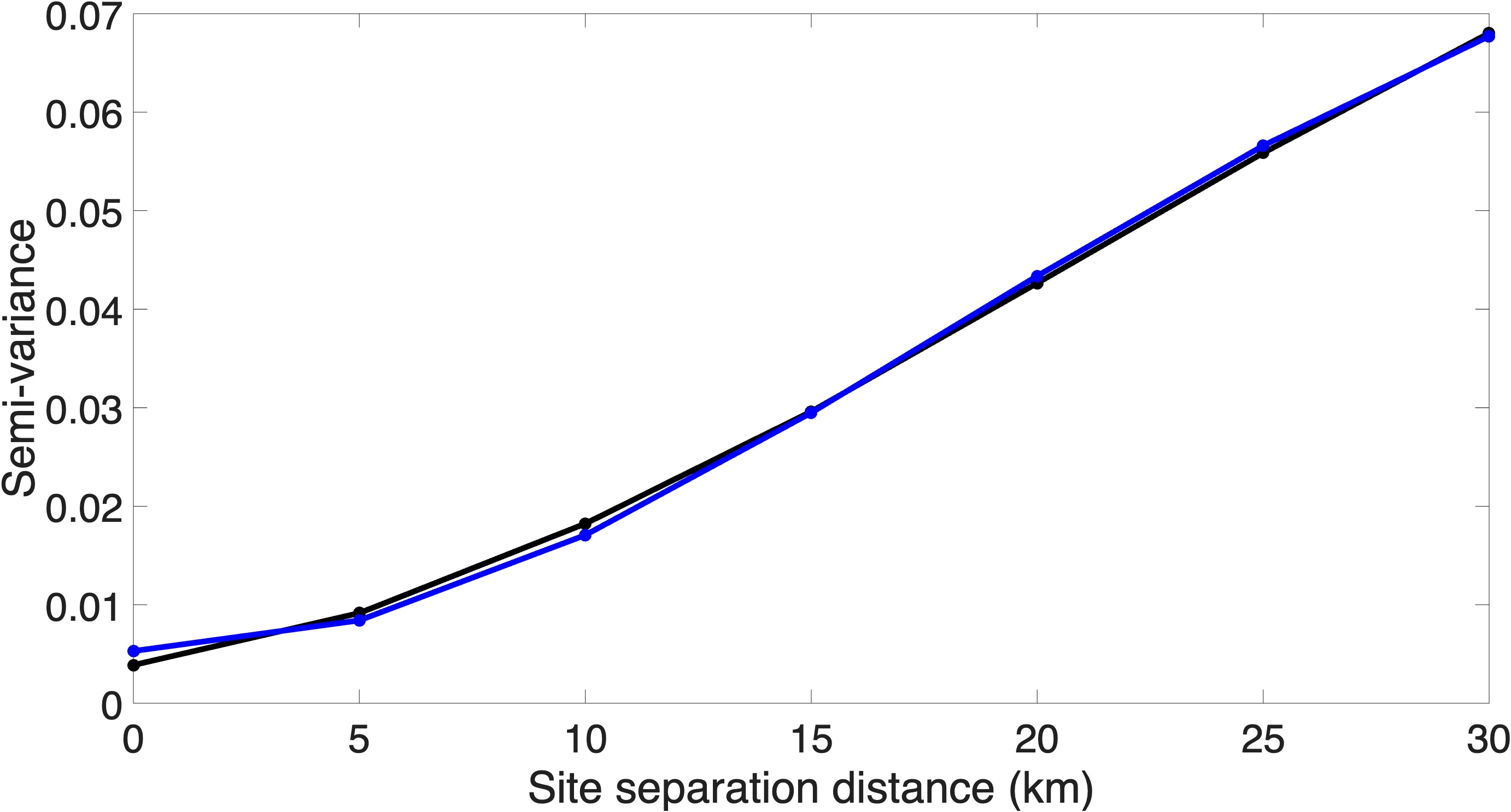}
        \caption{}
        \label{fig:CVAE_Semi_Variograms_Berkeley}
    \end{subfigure} 
     \begin{subfigure}{.45\linewidth}
        \centering
        \includegraphics[width=\linewidth]{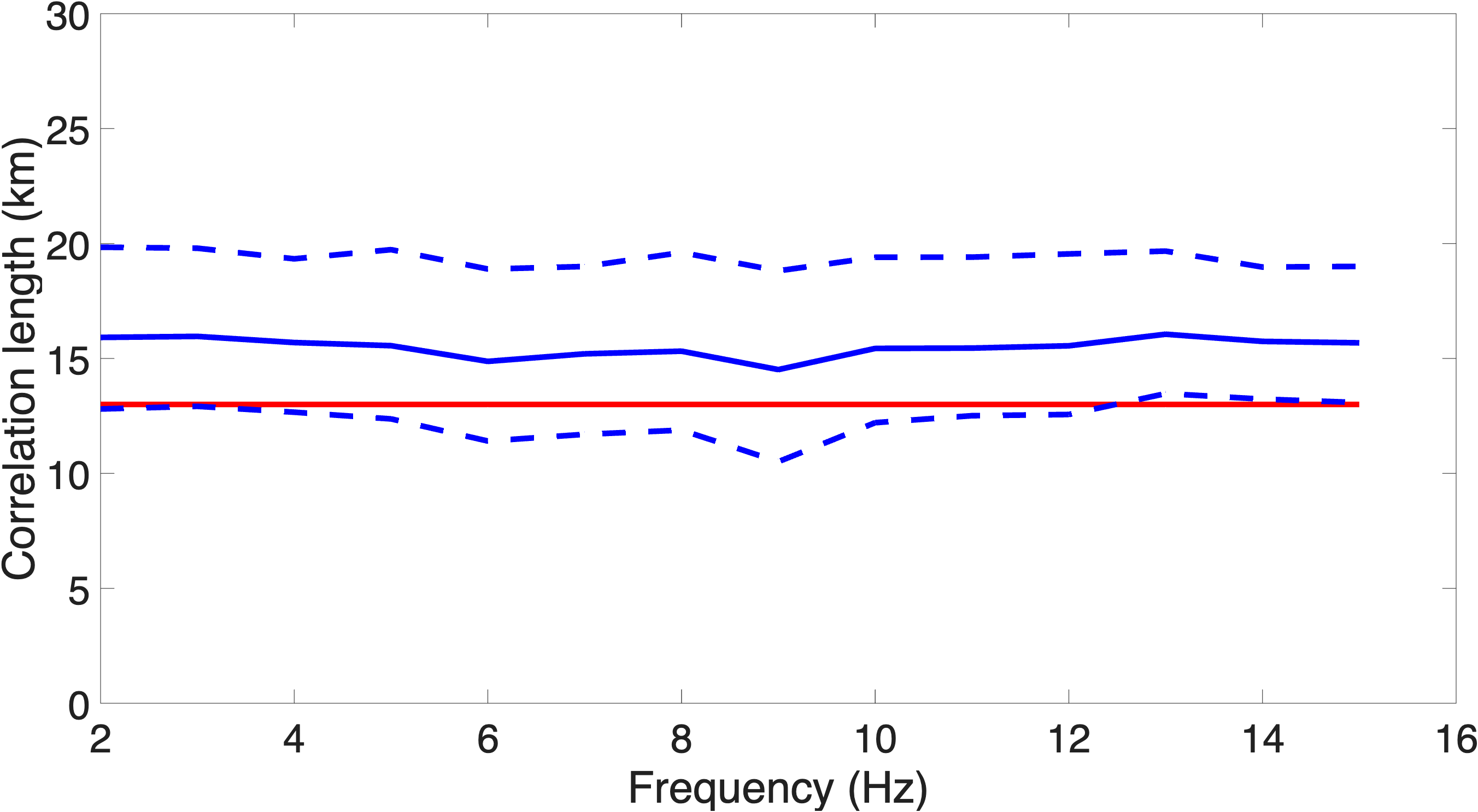}
        \caption{}
        \label{fig:validation_correlation_lengths}
    \end{subfigure}
    \caption{Spatial correlation lengths of predicted path effects.  (a) Semivariogram of the CGM-FAS prediction (black) and fitted squared-exponential correlation function (blue) for the Berkeley event at 10 Hz as a function of site separation distance. The fitted correlation length is equal to 18 km. 
    (b) Frequency dependence of correlation lengths for all events. The red line shows LANN25 from \cite{Lacour2025b}. The blue lines show correlation lengths from CGM-FAS predictions for each available earthquake in the dataset. The solid line represents the mean correlation length, while the dashed lines show the standard deviation.}
    \label{fig:spatial_correlation_validation}
\end{figure}

CGM-FAS produces smoother spatial patterns than LANN25. The LANN25 model exhibits strong azimuthal variation, particularly near the source, due to its prescribed azimuthal correlation function. This results in similar path effect predictions across different distances within the same azimuthal direction, which can lead to spatial prediction inaccuracies.
CGM-FAS allows variations path effects at sites within the same azimuthal direction but at different distances from the event. 
To quantify the spatial smoothness and continuity of the predictions, we examined the spatial correlation lengths of the predicted path terms from CGM-FAS. We computed semi-variograms of the predicted path terms at 10 Hz for the Berkeley event shown in Figure \ref{fig:Maps_Path_Pred_GP_VAE} as a function of site separation distance and fitted a squared-exponential correlation function following \cite{Lacour2025b}. The semivariogram shown in Figure \ref{fig:CVAE_Semi_Variograms_Berkeley} for 10 Hz  approaches zero at small distances with a good fit to the squared-exponential function with the correlation length of 18 km, suggesting strong continuity in the predictions. 

We extended the analysis to the entire frequency range and for all the events in the dataset. We show the mean and standard deviation of the estimated correlation lengths in Figure~\ref{fig:validation_correlation_lengths}. CGM-FAS exhibits correlation lengths with values ranging from approximately 13 to 20km, and a mean of approximately 16 km, slightly larger than the LANN25 value of 14 km. While LANN25 provides a single correlation length across station locations per event and per frequency, CGM-FAS learns correlation structures that vary across earthquake locations and frequencies. This variability allows more flexibility in path effect predictions than LANN25, which prescribes a fixed correlation length for all station and earthquake locations.

\subsection*{Interfrequency Correlation}

Accurate modeling of interfrequency correlation is critical for capturing the variability of structural response in seismic fragility and risk studies. \cite{Bayless2018b} showed that without interfrequency correlation, the variability in structural response may be underestimated, leading to structural fragilities that are too steep and resulting in nonconservative estimates of seismic risk.  

\begin{figure}[H]
    \centering
    \begin{subfigure}[b]{0.45\linewidth}
        \centering
        \includegraphics[width=\linewidth]{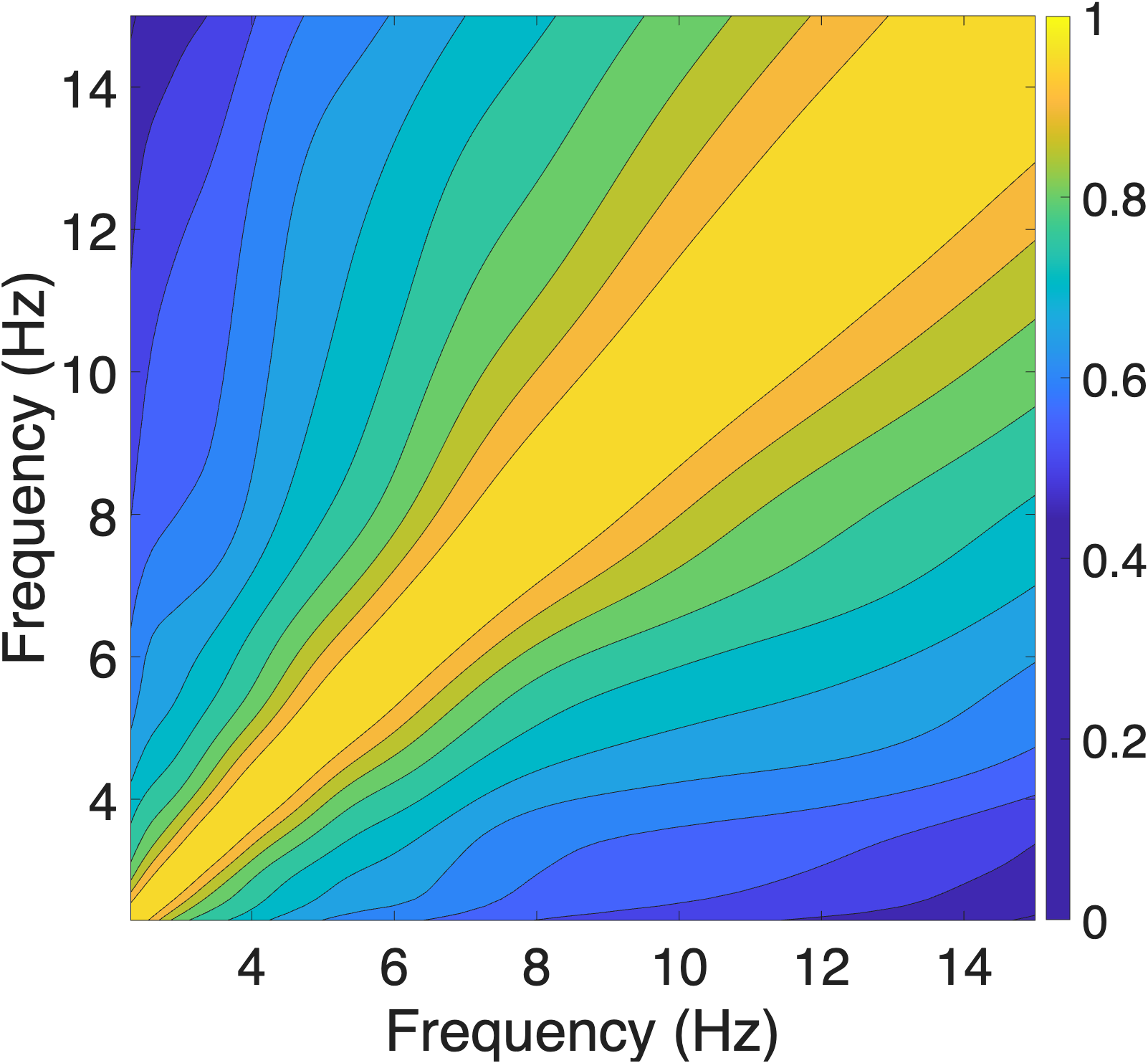}
        \caption{}
        \label{fig:Interfrequency_Correlation_Empirical}
    \end{subfigure}
    \hfill
    \begin{subfigure}[b]{0.45\linewidth}
        \centering
        \includegraphics[width=\linewidth]{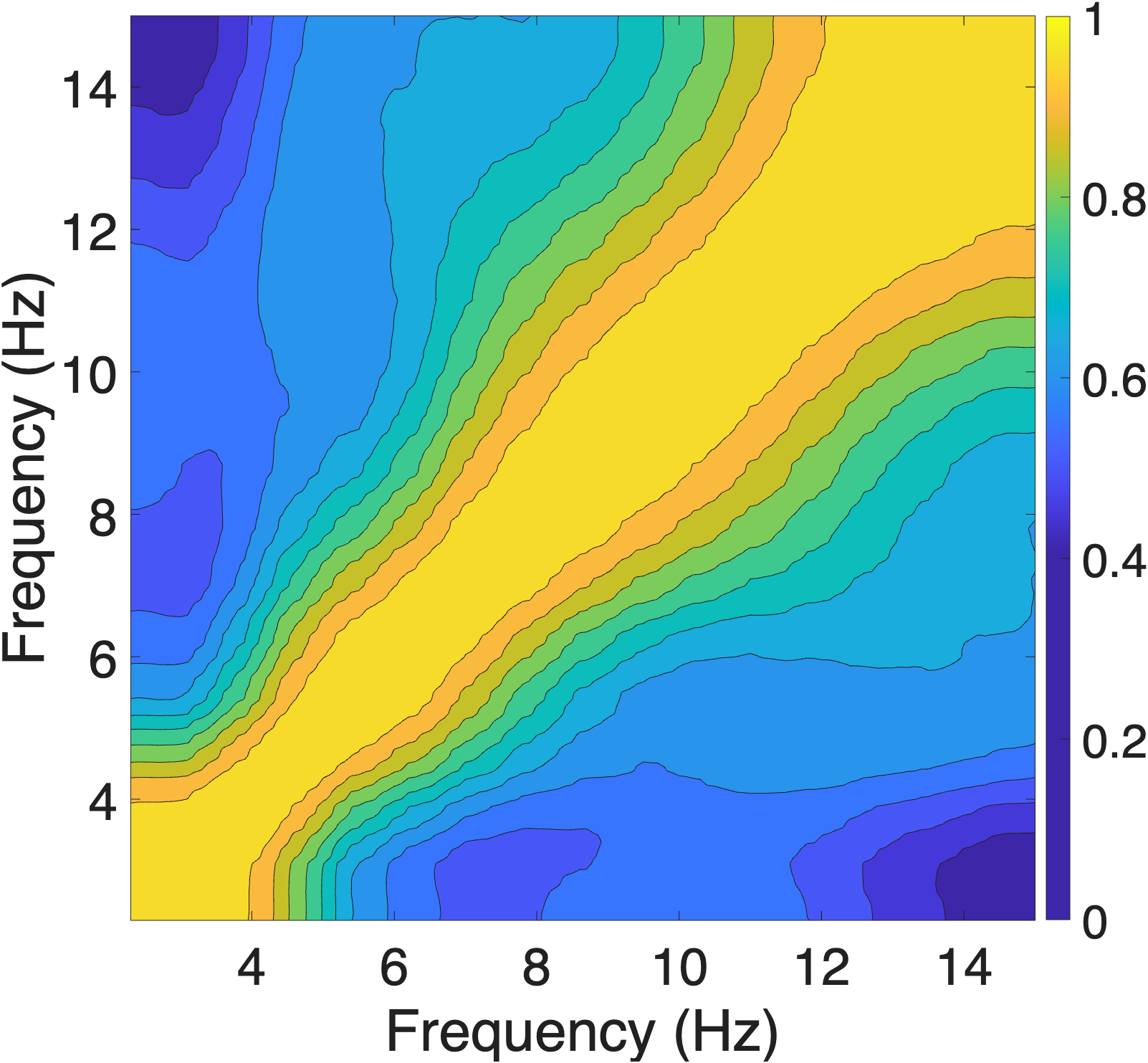}
        \caption{}
                \label{fig:Interfrequency_Correlation_Reconstructed}
    \end{subfigure}
\caption{Interfrequency correlation from (a) observations  and (b) CGM-FAS.}
 \label{fig:Interfrequency_Correlation}
\end{figure}

CGM-FAS learns across all frequencies simultaneously, while LANN25 models each frequency independently. To assess the CGM-FAS model's ability to capture dependencies between different frequency components, we computed the interfrequency correlation matrices from both the actual within-site residuals and the CGM-FAS reconstructions. 
Figure \ref{fig:Interfrequency_Correlation_Empirical} shows the empirical interfrequency correlation structure from the dataset, which was obtained by calculating the correlation matrix between the observed within-site residuals across frequencies. The correlation is equal to 1 on the diagonal and decreases smoothly with increasing frequency separation. This decay is broader at higher frequencies than at lower frequencies, indicating that high-frequency components have stronger correlations over wider frequency ranges. This structure is consistent with observations in ground motion studies \citep{Bayless2018b}. Figure \ref{fig:Interfrequency_Correlation_Reconstructed} shows that the CGM-FAS model successfully reproduces this interfrequency correlation structure. The model captures the overall correlation structure and the broader decay of correlation at higher frequencies. However, some discrepancies are observed at the edges of the frequency domain, which can be attributed to boundary effects introduced by the sliding window approach used in the convolutions. These edge effects disappear between 4 Hz and 13 Hz, and we consider that CGM-FAS is reasonable within this range.


\section*{Discussion} \label{section:Discussion}

CGM-FAS differs from the GP framework in several fundamental ways. Instead of prescribing spatial continuity through predefined correlation functions as in GP models, CGM-FAS learns spatial dependencies directly from data. We demonstrated that CGM-FAS predictions were smooth and stable, and was able to to mitigate azimuthal artefacts in LANN25 arising from issues in the functional forms. 
Although we anticipate that CGM-FAS captures spatially varying correlation lengths, which is not currently not implemented in non-ergodic models using GP, such as LANN25 from \cite{Lacour2025b} or \cite{Lavrentiadis2023eas}, this feature needs to be further investigated.

While GP-based predictions are typically made independently at each frequency, we showed that CGM-FAS can learn and predict across multiple frequencies simultaneously, capturing interfrequency correlations. The posterior distribution is not assumed to be Gaussian, as CGM-FAS defines a nonlinear mapping between the latent variables and the data space through the encoder–decoder architecture. CGM-FAS does not directly provide statistical measures such as the mean or standard deviation; these must be inferred from ensembles of data realizations.

CGM-FAS offers significant computational and memory advantages over GP methods. Once trained, CGM-FAS generates predictions through passes in the neural network involving basic linear algebra operations, with cost scaling only with the number of prediction locations. In contrast, GP methods must solve large linear systems for each prediction, with computational cost and memory requirements that increase much more with the number of training and prediction locations. For example, predicting path effects over a $100 \times 100$ grid across 911 frequencies took approximately 10 seconds with CGM-FAS compared to an estimated 10 hours with standard GP methods. Efficient techniques like \citep{Lacour2022} and \citep{KuehnINLA} can reduce GP computational requirements but still require significantly more resources than CGM-FAS.

ML models for ground motions must balance fitting observations with preserving variability in the predictions. This contrasts with many ML applications in geophysics, which typically emphasize data fit. We demonstrate in Supplementary Section \ref{section:Supplement_Alpha} that this balance is achievable by tuning ML architectures using empirical guidance and by assuming that the standard deviation of the within-path residuals from CGM-FAS reflects the aleatory variability reported in previous GP-based studies. Future work will evaluate whether different ML approaches reproduce comparable levels of aleatory variability.

For non-ergodic path predictions to be used in PSHA, both the aleatory variability of the within-path residuals and the epistemic uncertainty in the path terms are necessary. CGM-FAS can estimate aleatory variability as a standard deviation of ensemble analysis of the within-path residuals. This allows the aleatory variability to vary spatially across different realizations, whereas GP methods typically assume a constant aleatory variability after accounting for the mean spatial trend. GP methods directly provide epistemic uncertainty in their predictions through the posterior covariance. However, CGM-FAS does not currently quantify epistemic uncertainty in the predicted path terms, which represents an important direction for future work. Two primary sources of epistemic uncertainty in CGM-FAS are data sparsity, due to limited or uneven observational coverage, and model-related uncertainty arising from architectural choices, hyperparameters, and optimization stochasticity. For example, station density is a prominent factor in influencing the performance of CGM-FAS, as we show in Supplementary Section \ref{section:Supplement_Epistemic}. \

While CGM-FAS uses CVAEs as its backbone, the framework could benefit from alternative architectures. Diffusion models, as demonstrated by CGM-Wave of \cite{Bi2025} for wavefield synthesis, may improve spatial resolution. The CGM framework can be extended to incorporate site and source effects by generating the spatially varying between-source and between-site residuals. Combining spatially varying source, site, and path effects at multiple frequencies is a promising direction for non-ergodic ground motion modeling, as it would remove the independent treatment of these effects that can be correlated.


\section*{Conclusions}

CGM-FAS successfully learns and predicts spatially varying non-ergodic path effects, producing maps comparable to those from traditional GP-based approaches such as LANN25, while exhibiting smoother spatial continuity. CGM-FAS addresses limitations of current GP methods for non-ergodic modeling. First, CGM-FAS learns spatial correlations directly from observations without prescribing correlation functions, allowing the model to capture complex spatial patterns that would be difficult to specify a priori in GP frameworks. Second, the computational efficiency of CGM-FAS enables prediction of path effects over finely discretized and broad geographic areas across frequencies in seconds, while the GP approach would require several hours for the same task if no additional numerical approximations are used. Third, CGM-FAS naturally captures interfrequency correlations across the entire frequency range, which are essential for accurate seismic risk assessments of structures sensitive to ground motion at multiple frequencies simultaneously.
Using the small earthquake dataset from San Francisco Bay Area, we demonstrated the performance of CGM-FAS. Future work includes quantifying epistemic uncertainty in CGM-FAS predictions for integration into seismic hazard applications, and extending the framework to model spatially varying source and site effects, ultimately providing a fully non-ergodic GMM that captures correlations between these effects over spatial and frequency domains.

\section*{Data and Resources}

The dataset used in this study was downloaded from the NCEDC database \citep{ncedc} and the preprocessed data is provided by \cite{Lacour2025c} at: https://www.designsafe-ci.org/data/browser/public/designsafe.storage.published/PRJ-4573.

\begin{figure}[H]
     \begin{subfigure}[c]{0.48\linewidth}
         \centering
         \includegraphics[width = 0.9\columnwidth]{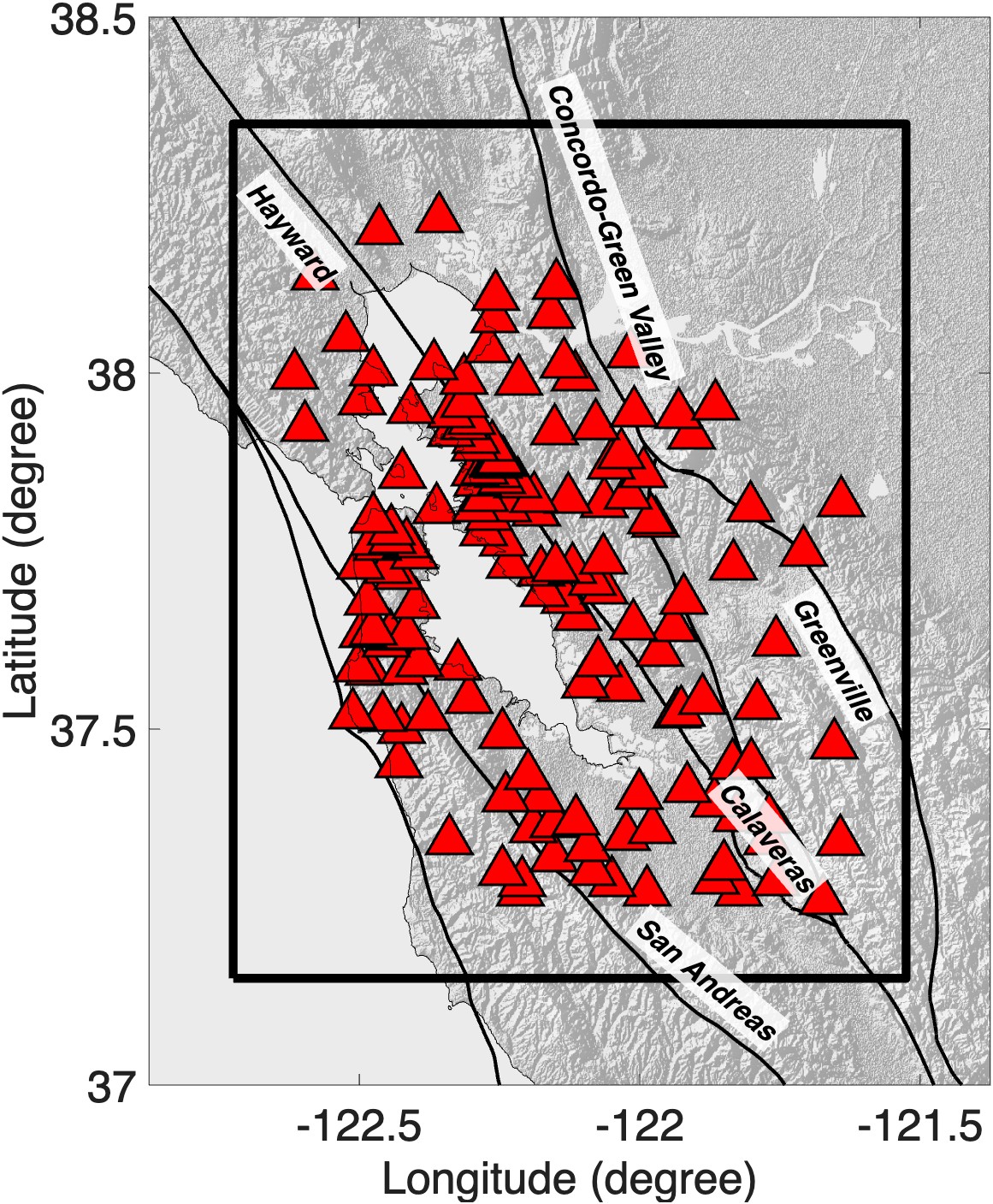} 
                  \caption{}
     \end{subfigure}  
     \hfill
     \begin{subfigure}[c]{0.48\linewidth}
         \centering
         \includegraphics[width = 0.9\columnwidth]{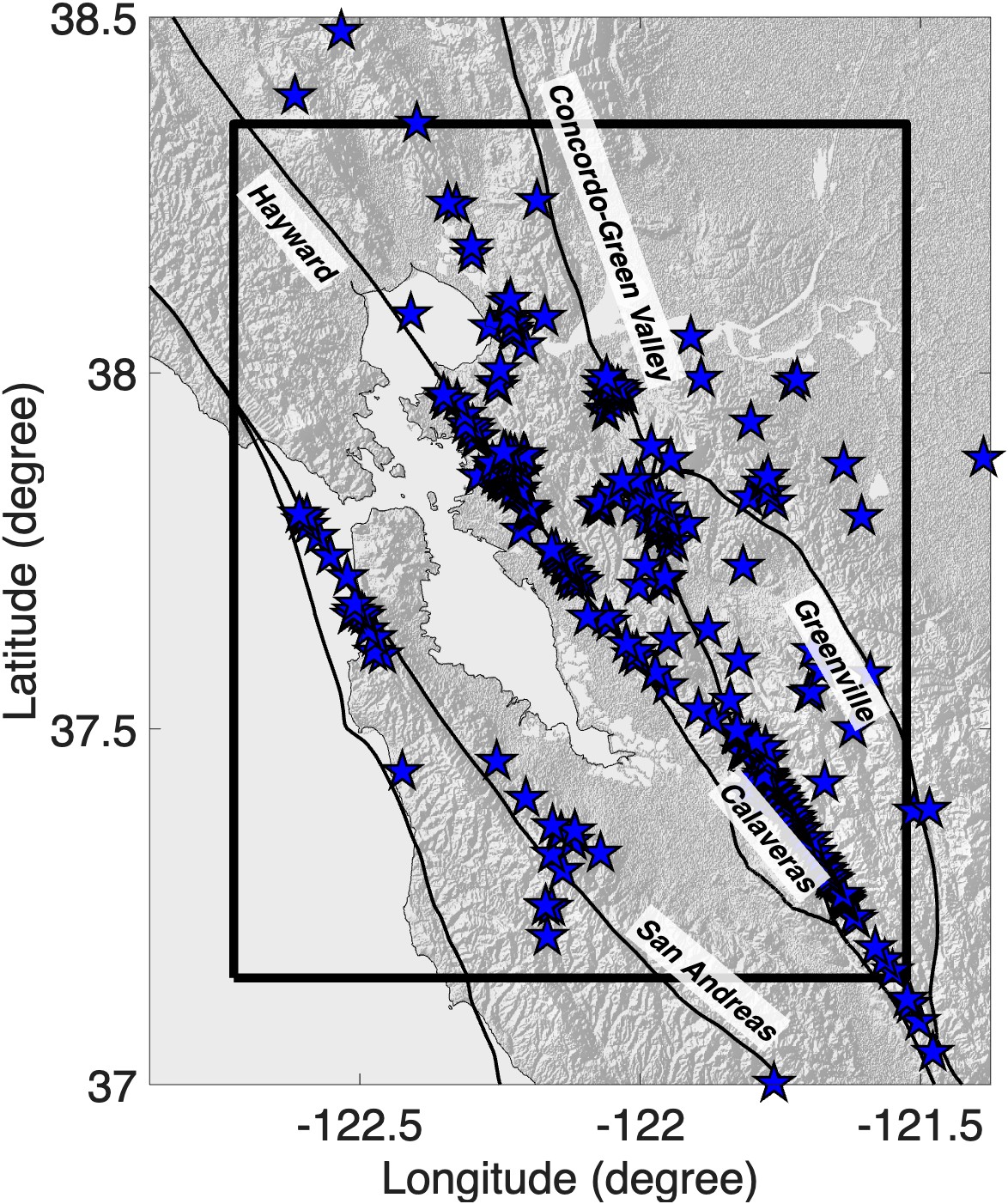}
                  \caption{}
     \end{subfigure}
         \caption{Station locations (a) and source locations (b) from the selected dataset from \citep{Lacour2025c} for the East-West component. Red triangles and blue stars show the selected station and source locations, respectively. The black box defines the spatial domain shown in later generated maps, which contains all selected stations and their associated recordings. The thin black lines indicate mapped faults in the USGS quartenary database. The earthquakes range between magnitude 1 and 4 and were recorded by broadband seismometers. More information on the dataset is available in \cite{Lacour2025b}.
         }
         \label{fig:Stations_and_Sources}
\end{figure}

The quaternary fault traces used in Figure \ref{fig:Stations_and_Sources} are obtained from the U.S. Geological Survey and New Mexico Bureau of Mines and Mineral Resources, Quaternary fault and fold database for the United States at: https://www.usgs.gov/natural-hazards/earthquake-hazards/faults.

The source code for CGM-FAS will be available upon acceptance at https://github.com/maxlacour/CGM-FAS. 

The supplementary material provides additional details on the calibration of the generation parameter $\alpha$, convergence analysis, spatial continuity validation, and epistemic uncertainty quantification for the CGM-FAS model.

\section*{Acknowledgments}

This work was supported by the Laboratory Directed Research and Development Program of Lawrence Berkeley National Laboratory under U.S. Department of Energy Contract No. DE-AC02-05CH11231. We also would like to acknowledge the Statewide California Earthquake Center (SCEC). Norman Abrahamson provided useful review comments on the draft report.


\bibliography{bibliography}

\clearpage
\newpage

\section*{Full Mailing Address for Corresponding Author}

\noindent Maxime Lacour \\
University of California, Berkeley\\
Department of Civil \& Environmental Engineering \\
750 Davis Hall \\
Berkeley, CA 94720 \\

\clearpage
\newpage

\section*{List of Tables and Figures}


%

\section*{Supplementary Materials}
\setcounter{figure}{0}
\renewcommand{\thefigure}{S\arabic{figure}}
\setcounter{table}{0}
\renewcommand{\thesection}{Text S\arabic{section}}
\renewcommand{\thetable}{S\arabic{table}}

\textbf{{Modeling Non-Ergodic Path Effects Using Conditional Generative Model for Fourier Amplitude Spectra}} \\

\noindent {\textbf{Maxime Lacour, Pu Ren, Rie Nakata, Nori Nakata, Michael Mahoney}} \\


\noindent \textbf{Description of the Supplementary Material}

\noindent This supplementary material provides additional details on the calibration of the generation parameter $\alpha$, convergence analysis, spatial continuity validation, and epistemic uncertainty quantification for the CGM-FAS model. \\

\noindent \textbf{Supplementary Text}

\noindent This supplementary material contains four text descriptions about:

\noindent \ref{section:Supplement_Alpha} describes the calibration procedure for the generation parameter $\alpha$ to balance reconstruction accuracy and latent space regularization in the CVAE loss function.

\noindent \ref{section:Supplement_Convergence} presents convergence analysis showing how the mean and standard deviation of CGM-FAS predictions stabilize with increasing sample size.

\noindent \ref{section:Supplement_Epistemic} describes the epistemic uncertainty quantification experiments using k-means clustering with varying station separation distances (5-50 km).

\noindent \textbf{List of Supplemental Table Captions} 

\noindent Table \ref{tab:sup_math_notation} Mathematical notation used throughout the manuscript. \\

\noindent \textbf{List of Supplemental Figure Captions} 

\noindent Figure \ref{fig:alpha_calibration} Calibration of the $\alpha$ parameter. Left column shows histograms of within-path residuals at 10 Hz comparing the target distribution from LANN25 (red) with CGM-FAS predictions (blue), with mean shown as solid lines and standard deviation as dashed lines. Right column shows the standard deviation as a function of frequency. Results shown for (a,b) $\alpha = 5\times10^{-7}$, (c,d) $\alpha = 3\times10^{-5}$, and (e,f) $\alpha = 3\times10^{-4}$. As $\alpha$ increases, the standard deviation of the CGM-FAS within-path residuals decreases, with optimal calibration achieved at $\alpha = 3\times10^{-5}$ where the model matches the theoretical target value $\phi_{\text{SP,NE}} = 0.40$. Higher values of $\alpha$ lead to overfitting, with standard deviations below the target.

\noindent Figure \ref{fig:Mean_Sd_Convergence} Convergence analysis of CGM-FAS predictions showing (a) predicted mean within-site residuals and (b) predicted standard deviation as a function of the number of generated samples.

\noindent Figure \ref{fig:spatial_extrapolation} Epistemic uncertainty quantification showing (a) illustration of five station subsets selected using k-means clustering with 20 km minimum separation, and (b) effects of station separation on MSE at 10 Hz. 

\clearpage
\newpage


\begin{table}[h!]
\caption{Summary of mathematical notations. "Dist." stands for "Distribution".}

\centering
\small
\setlength{\tabcolsep}{4pt}
\begin{tabular}{ccl}
\hline
Symbol & Type & Description \\
\hline
$FAS_{NE}$, $FAS_{E}$ & - & Non-ergodic, ergodic velocity FAS \\
$M$, $R_{Rup}$, $Z_{hyp}$ & Scalar & Magnitude, rupture distance, hypocenter depth \\
$te_e$ & Scalar & Earthquake e coordinates (lat, lon, depth)  \\
$ts_s$ & Scalar & Stations s coordinates (lat, lon)  \\
$\mathbf{te}$, $\mathbf{ts}$ & Vector & Earthquake and station coordinate vectors \\
$\delta L2L_e(te_e)$ & Scalar & Mean source term deviation from ergodic model \\
$\delta S2S_s(ts_s)$ & Scalar & Mean site term deviation from ergodic model \\
$\delta P2P_{es}(te_e, ts_s)$ & Scalar & Mean path term deviation from ergodic model \\
$\delta B_e^0$ & Scalar & Aleatory source residuals \\
$\dWS$ & Vector & Within-site residuals across sources and stations \\
$\dWSobs$ & Vector & Observed within-site residuals \\
$\dWSpred$ & Vector & Predicted within-site residuals \\
$\hat{\mathbf{\dWS_{obs}}}$ & Vector & Reconstructed within-site residual by CGM-FAS \\
$\delta WSP_{es}$ & Scalar & Within-path residuals \\
$\mathbf{z}$ & Vector & Latent variables \\
$p(\mathbf{z})$ & Dist. & Prior distribution of latent variables \\
$\boldsymbol{\mathbf{\mu}}, \boldsymbol{\mathbf{\sigma}}$ & Vector & mean and standard deviation of latent variable \\
$p_{\boldsymbol{\theta_{\rencoder}}(\mathbf{z})}$ & Dist. & Distribution of the encoder \\
$p_{\boldsymbol{\theta_{\rdecoder}}(\mathbf{z})}$ & Dist. & Distribution of the decoder \\
$\Koo$, $\Kpp$, $\Kop$ & Matrix & GP covariance matrices   \\
$ \boldsymbol{\dPtoP}_{\boldsymbol{\mathrm{pred}},\vtheta_{\boldsymbol{\mathrm{GP}}}}$ & Vector & GP mean path prediction  \\
$\delta P2P_\mathbf{{\mathrm{pred},VAE}}({te}_{e},{ts}_s)$ & Vector & CGM-FAS mean path prediction \\
$ \boldsymbol{\psi}_{\boldsymbol{\mathrm{pred}},\vtheta_{\boldsymbol{\mathrm{GP}}}} $ & Vector & GP epistemic uncertainty  \\
$\Delta R_{rup}$ & Scalar & Difference in rupture distances \\
$\Delta Az$ & Scalar & Difference in azimuths \\
$|SS'|$ & Scalar & Distance between site locations \\
$\phi_{SP,NE}$ & Scalar & Standard deviation of aleatory within-path residuals\\
$\phi_{P2P}$ & Scalar & Standard deviation of GP spatially correlated path effects \ \\
$\rho_R$, $\rho_{Az}$, $\rho_S$ & Scalar & GP hyperparameters of path correlation function \\
$\boldsymbol{\theta_{\rencoder}},\boldsymbol{\theta_{\rdecoder}}$ & Vector & Decoder and encoder network hyperparameters \\
$\boldsymbol{\theta_{GP}}$ & Vector & GP hyperparameters ($\phi_{P2P}$, $\phi_{SP,NE}$, $\rho_R$, $\rho_{Az}$, $\rho_S$) \\
$k_{PP'}$ & Scalar & Covariance between paths P and P' \\
$\delta(P, P')$ & Scalar & Kronecker delta function (1 if paths identical, 0 otherwise) \\
$\alpha$ & Scalar & Reconstruction-KL trade-off parameter  \\
$\mathcal{L}(\boldsymbol{\theta}_{\mathrm{encoder}},\boldsymbol{\theta}_{\mathrm{decoder}};\boldsymbol{\delta WS}_{\mathrm{obs}})$ & Scalar & Loss function for CGM-FAS \\
$\mathbf{I} $ & Matrix & Identity matrix \\
\hline
\end{tabular}
\label{tab:sup_math_notation}
\end{table}

\section{Effects of the Trade-off Parameter $\alpha$} \label{section:Supplement_Alpha}

The hyperparameter $\alpha$ in the CVAE loss function Equation \eqref{eq:loglik_CGMFAS} controls the balance between reconstruction accuracy (MSE term) and latent space regularization (KL divergence term). This trade-off directly affects the variability of generated samples around the predicted path terms. We calibrated $\alpha$ to ensure that the range of the generated within-site residuals is consistent with the aleatory variability $\phi_{SP, NE}$ of the within-path residuals that remains after accounting for spatial correlation in path effects. Setting $\alpha$ too low causes the model to overfit and artificially over-explain the observations, producing insufficient variability. Conversely, setting $\alpha$ too high results in excessive variability that exceeds physically realistic generations. The optimal value ensures CGM-FAS generates realistic samples with variability consistent with what can be explained through spatial correlation alone.

\begin{figure}[H]
    \centering
    \begin{subfigure}[t]{0.45\linewidth}
        \centering
        \includegraphics[width=\linewidth]{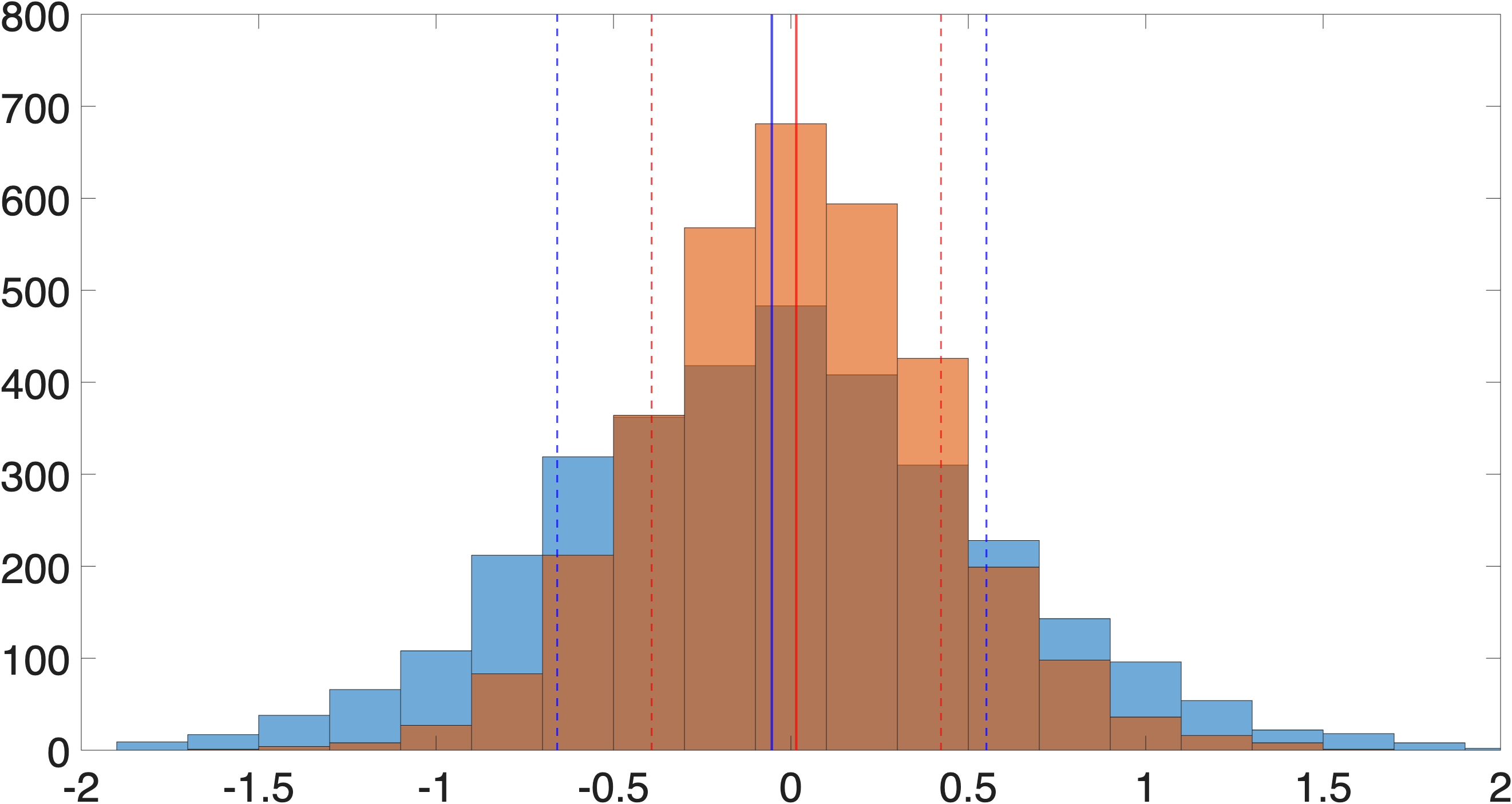}
        \caption{}
        \label{fig:alpha_1}
    \end{subfigure}
    \hfill
    \begin{subfigure}[t]{0.47\linewidth}
        \centering
        \raisebox{-0.85em}{\includegraphics[width=\linewidth]{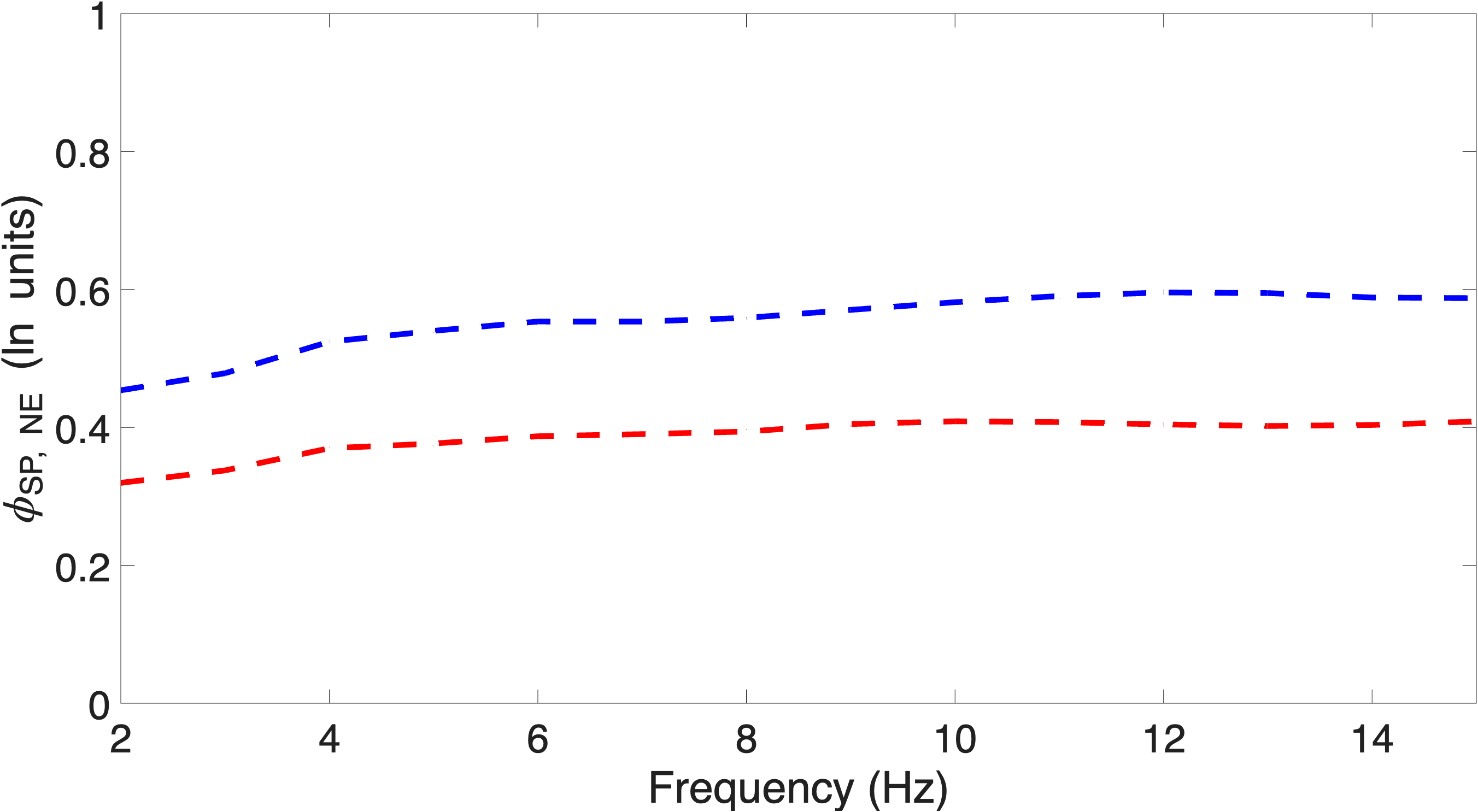}}
        \caption{}
        \label{fig:alpha_stddev_1}
    \end{subfigure}

    \vspace{0.5em}

    \begin{subfigure}[t]{0.45\linewidth}
        \centering
        \includegraphics[width=\linewidth]{Histogram_Alpha_x03.png}
        \caption{}
        \label{fig:alpha_2}
    \end{subfigure}
    \hfill
    \begin{subfigure}[t]{0.47\linewidth}
        \centering
        \raisebox{-0.85em}{\includegraphics[width=\linewidth]{Sd_WSP_GP_CVAE.png}}
        \caption{}
        \label{fig:alpha_stddev_2}
    \end{subfigure}

    \vspace{0.5em}

    \begin{subfigure}[t]{0.45\linewidth}
        \centering
        \includegraphics[width=\linewidth]{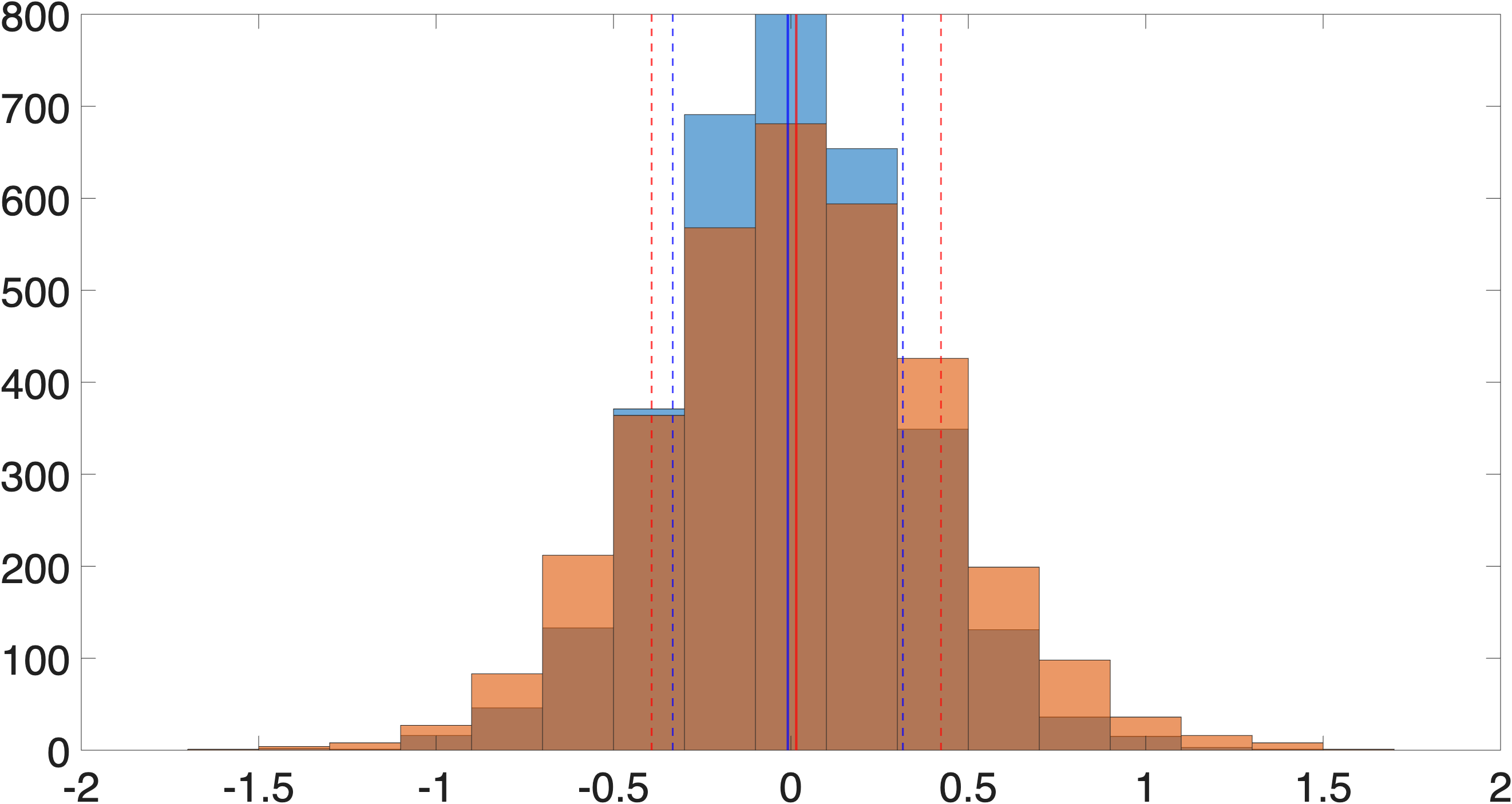}
        \caption{}
        \label{fig:alpha_3}
    \end{subfigure}
    \hfill
    \begin{subfigure}[t]{0.47\linewidth}
        \centering
        \raisebox{-0.85em}{\includegraphics[width=\linewidth]{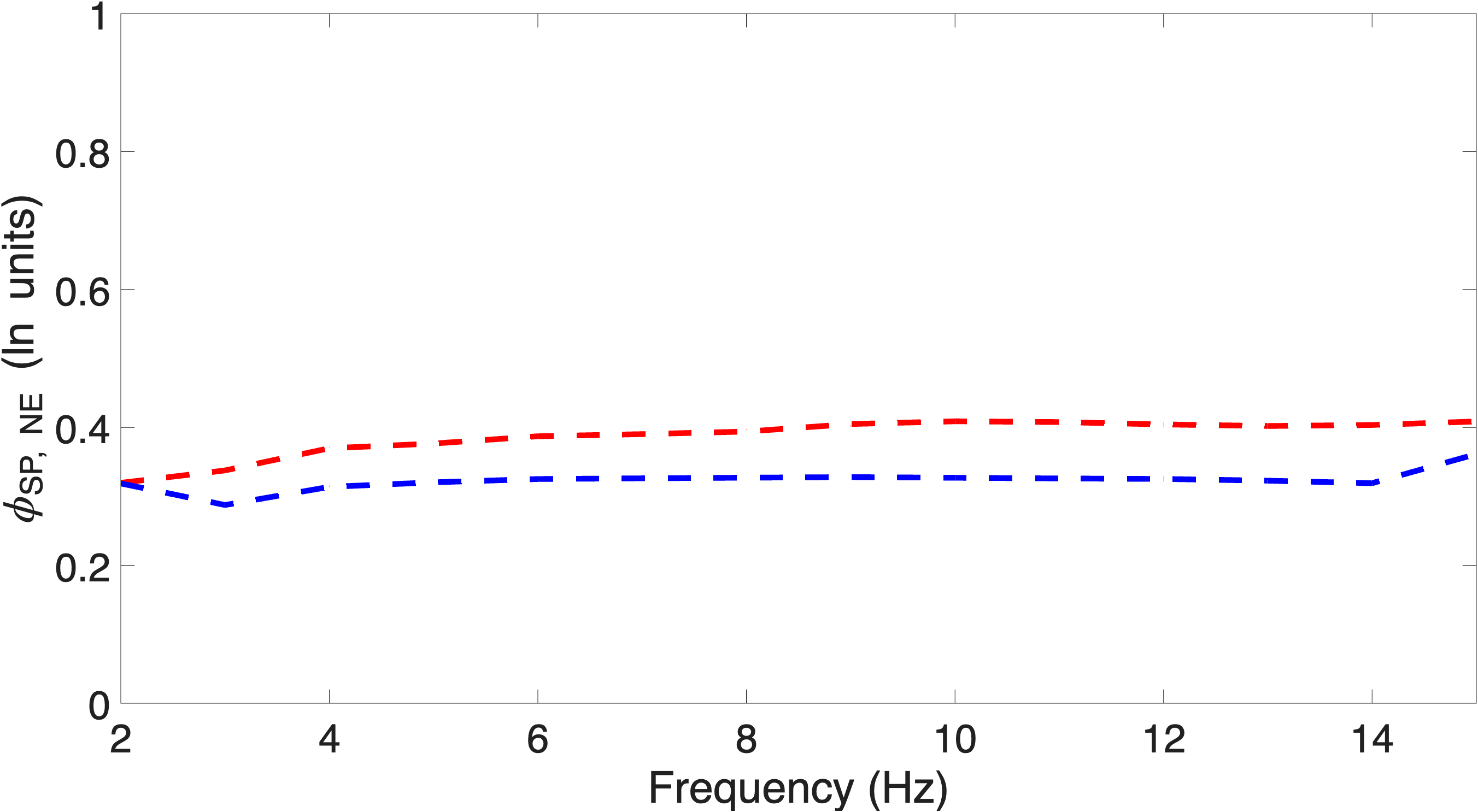}}
        \caption{}
        \label{fig:alpha_stddev_3}
    \end{subfigure}
    
    \caption{Calibration of the $\alpha$ parameter. Left column shows histograms of within-path residuals at 10 Hz comparing the target distribution from LANN25 (red) with CGM-FAS predictions (blue). Right column shows the standard deviation as a function of frequency. The mean is shown with solid lines and standard deviation with dashed lines in the histograms. Results shown for (a,b) $\alpha = 5\times10^{-7}$, (c,d) $\alpha = 3\times10^{-5}$, and (e,f) $\alpha = 3\times10^{-4}$. As $\alpha$ increases, the standard deviation of the CGM-FAS within-path residuals decreases, with optimal calibration achieved at $\alpha = 3\times10^{-5}$ where the model matches the theoretical target value $\phi_{\text{SP,NE}} = 0.40$. }
    \label{fig:alpha_calibration}
\end{figure}

We assess the calibration of the $\alpha$ parameter by comparing the distribution of within-path residuals from CGM-FAS against the target distribution from LANN25 (Figure \ref{fig:alpha_calibration}). The left column shows histograms of within-path residuals at 10 Hz, while the right column shows the standard deviation as a function of frequency across the 2-15 Hz range. Results are shown for three values of $\alpha$: (a,b) $\alpha = 5\times10^{-7}$, (c,d) $\alpha = 3\times10^{-5}$, and (e,f) $\alpha = 3\times10^{-4}$. Each histogram compares CGM-FAS predictions (blue) with the target LANN25 distribution (red), with mean values shown as solid lines and standard deviations shown as dashed lines. At $\alpha = 5\times10^{-7}$ (Figures \ref{fig:alpha_1} and \ref{fig:alpha_stddev_1}), the standard deviations remain consistently above the target value $\phi_{\text{SP,NE}} = 0.40$ across all frequencies. As $\alpha$ increases to $3\times10^{-5}$ (Figures \ref{fig:alpha_2} and \ref{fig:alpha_stddev_2}), the standard deviation converges to the target value across the frequency range, with the CGM-FAS distribution closely matching both the mean and standard deviation of the target distribution. At $\alpha = 3\times10^{-4}$ (Figures \ref{fig:alpha_3} and \ref{fig:alpha_stddev_3}), the standard deviations fall below the target value.

\section{Convergence of CGM-FAS Prediction Statistics} \label{section:Supplement_Convergence}

\begin{figure}[H]
\centering
\begin{subfigure}{0.48\linewidth}
    \centering
    \includegraphics[width=0.9\columnwidth]{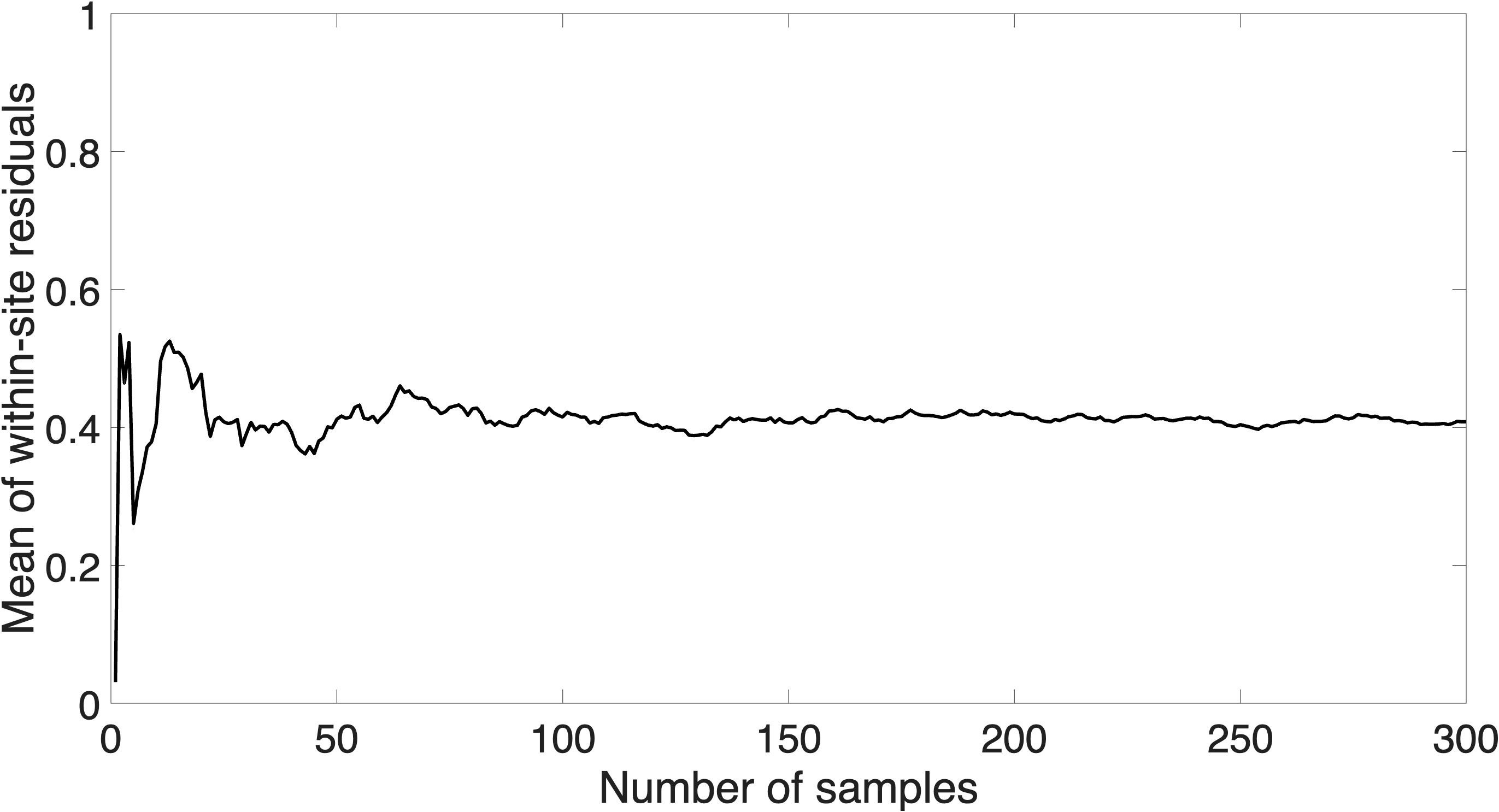}
    \caption{}
    \label{fig:Error_Mean}
\end{subfigure} 
\hfill
\begin{subfigure}{0.48\linewidth}
    \centering
    \includegraphics[width=0.9\columnwidth]{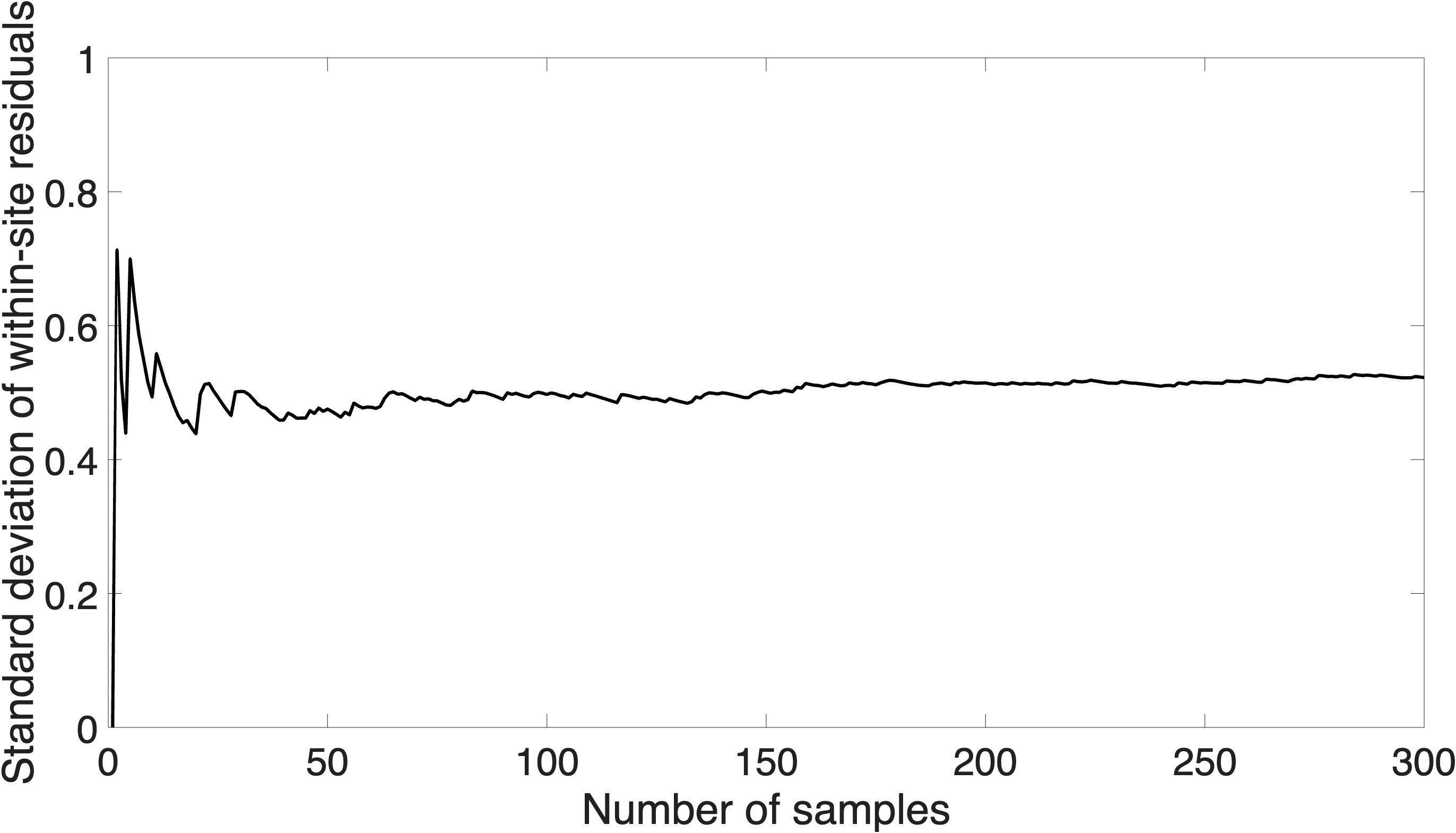}
    \caption{}
    \label{fig:Error_Sd}
\end{subfigure}

\caption{Convergence analysis of CGM-FAS predictions with increasing sample size. (a) Predicted mean within-site residuals and (b) predicted standard deviation for one source-site pair as a function of the number of generated samples. Both statistics stabilize after approximately 100 samples.}
\label{fig:Mean_Sd_Convergence}
\end{figure}

To assess the convergence behavior of CGM-FAS predictions, we examine how the mean and standard deviation of within-site residuals stabilize as more samples are generated. We select one source-site pair from the dataset with coordinates (37.3932, -121.7810) and (37.3416,  -121.6426) respectively, which are separated by 15 km from the dataset and compute the mean and standard deviation using sample sizes ranging from 1 to 300. Figure \ref{fig:Mean_Sd_Convergence} displays the mean and standard deviation of the within-site residuals with the number of generated samples. Both statistics start to stabilize beyond approximately 100 samples, and we use 200 samples for all predictions to compute the path effects.

\section{Effects of Station Density} \label{section:Supplement_Epistemic}

\begin{figure}[H]
\centering
\begin{subfigure}{0.32\linewidth}
    \centering
    \includegraphics[width=\columnwidth]{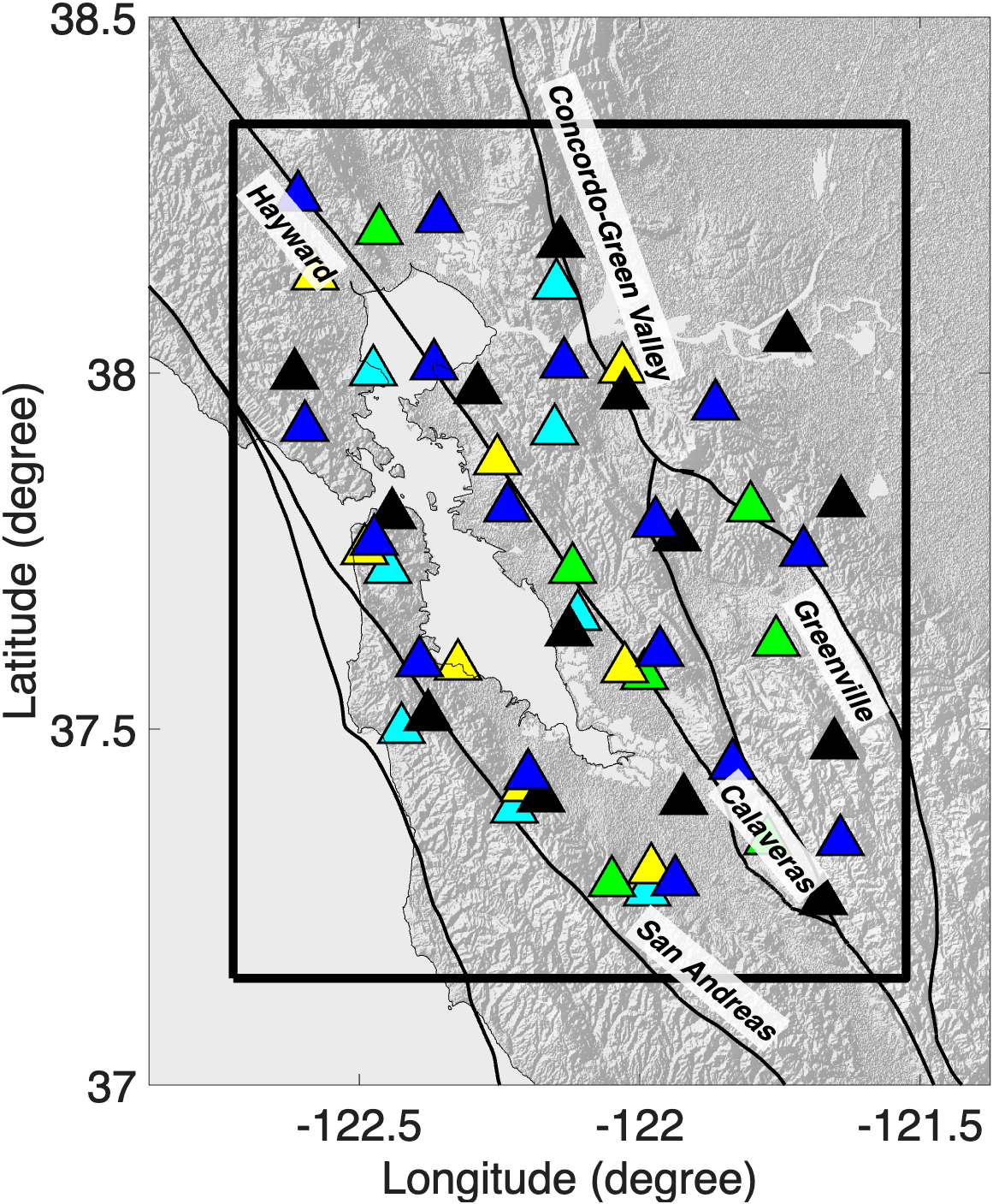}
    \caption{}
    \label{fig:Subset_Stations}
\end{subfigure} 
\begin{subfigure}{0.52\linewidth}
    \centering
    \includegraphics[width=\columnwidth]{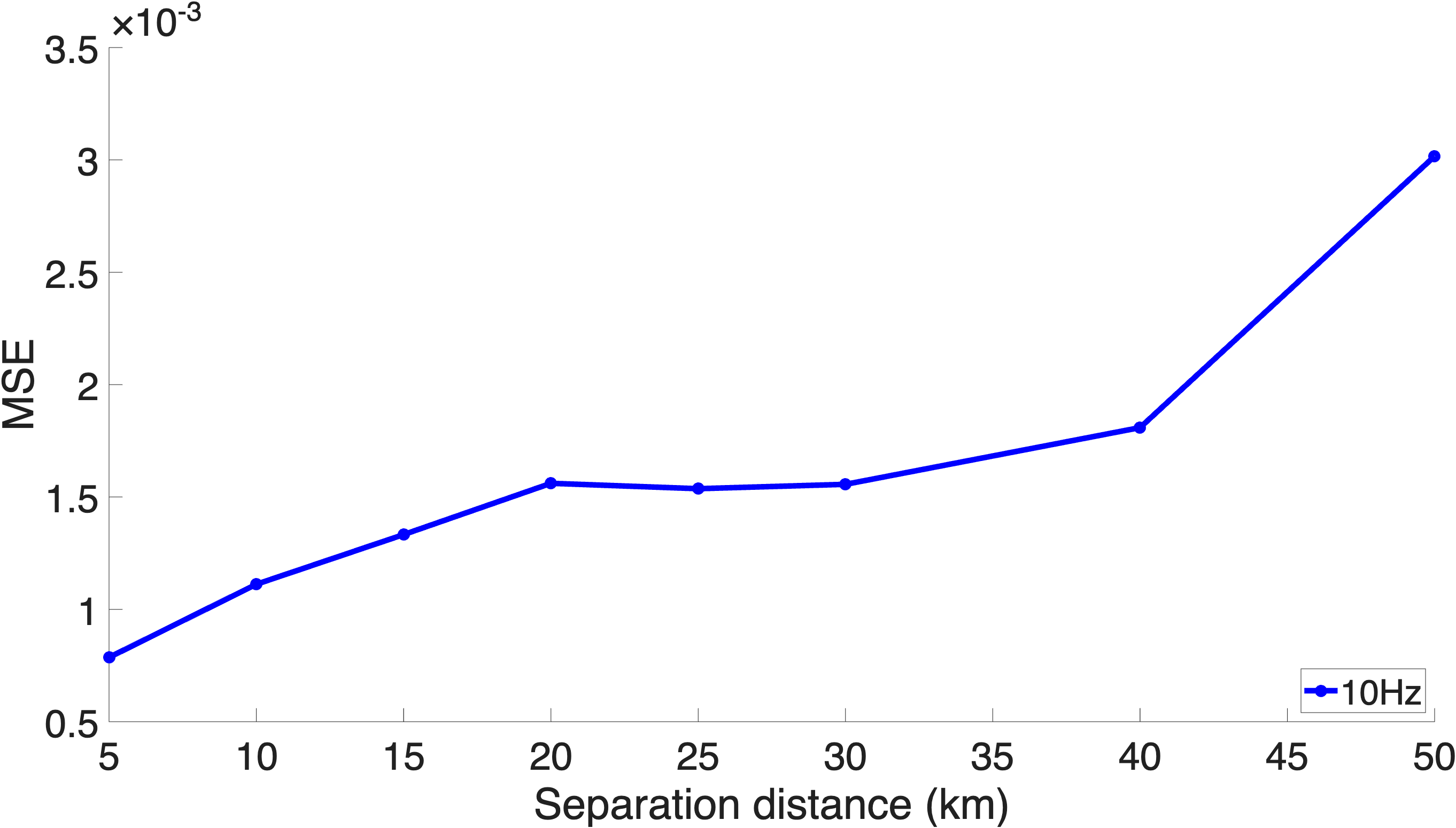}
    \caption{}
    \label{fig:Subsets_MSE}
\end{subfigure}

\caption{Effects of station density. (a) Illustration of the 5 station subsets selected using k-means clustering with 20 km minimum separation, each color representing a subset.  (b) Effects of station separation to mean-squared reconstruction error at 10 Hz.}
\label{fig:spatial_extrapolation}
\end{figure}

In order to assess the effects of sparse spatial coverage of sensors, we evaluated CGM-FAS performance as a function of training data density. We systematically reduced the available station coverage by selecting subsets of stations using k-means clustering, with minimum separation distances between training stations varying from 5 km to 50 km. Each experiment was repeated with five independent random subsets to ensure robust statistics. Figure \ref{fig:Subset_Stations} shows an example of five randomly selected subsets with 20 km minimum separation distance. We trained CGM-FAS using the same architecture as the full dataset, then predicted path effects at all source-site pairs from the complete dataset. We quantified prediction accuracy by calculating the MSE between observed and predicted within-site residuals at the held-out locations, averaged over the five random subsets. Figure \ref{fig:Subsets_MSE} shows how this spatial extrapolation error increases with station separation distance at 10 Hz. Minor fluctuations in the MSE curves occur because each k-means clustering produces a slightly different set of stations. Some sets of stations are better positioned to capture ground motion patterns in certain areas, resulting in better model performance. 

\end{document}